\documentclass[lettersize,journal]{IEEEtran}
\usepackage{amsmath,amsfonts}

\usepackage{subfigure}
\usepackage{array}
\usepackage[caption=false,font=footnotesize,labelfont=rm,textfont=rm]{subfig}
\usepackage{textcomp}
\usepackage{stfloats}
\usepackage{url}
\usepackage{verbatim}
\usepackage{graphicx}
\usepackage{subfig}
\usepackage{cite}
\usepackage{graphicx}
\usepackage{amsmath}
\usepackage{amssymb}
\usepackage{booktabs}
\usepackage{multirow}
\usepackage{bbding}
\usepackage{xcolor}
\usepackage{array}
\usepackage[numbers]{natbib}
\usepackage{threeparttable}

\usepackage[ruled,linesnumbered]{algorithm2e}
\definecolor{redcolor}{rgb}{0.761, 0.328, 0.263}
\definecolor{bluecolor}{rgb}{0.192, 0.545, 0.886}

\definecolor{darkred2}{rgb}{0.7, 0.0, 0.0}

\begin{document}
	
	\title{Hierarchical Vision-Language Interaction for Facial Action Unit Detection}
	
	\author{Yong Li,
		Yi Ren,
		Yizhe Zhang,
		Wenhua Zhang,
		Tianyi Zhang,
		Muyun Jiang,
		Guo-Sen Xie,
		Cuntai Guan ~\IEEEmembership{Fellow,~IEEE}
		\thanks{Corresponding author: Cuntai Guan, Guo-Sen Xie.}
		\thanks{Yong Li is with the School of Computer Science and Engineering, and the Key Laboratory of New Generation Artificial Intelligence Technology and Its Interdisciplinary Applications, Southeast University, Nanjing 210096, China. E-mail: mysee1989@gmail.com.}
		\thanks{Tianyi Zhang is with Key Laboratory of Child Development and Learning Science (Ministry of Education), School of Biological Sciences and Medical Engineering, Southeast University, Nanjing, China. E-mail: t.zhang@seu.edu.cn.}
		\thanks{Yi Ren, Yizhe Zhang, Wenhua Zhang and Guo-Sen Xie are with the School of Computer Science and Engineering, Nanjing University of Science and Technology, Nanjing, 210094, China. E-mail: yi.ren@njust.edu.cn, zhangyizhe@njust.edu.cn, whzhang@njust.edu.cn, gsxiehm@gmail.com.}
		\thanks{Cuntai Guan and Muyun Jiang are with the School of Computer Science and Engineering, Nanyang Technological University, 50 Nanyang Avenue, Singapore, 639798. E-mail: (ctguan, james.jiang)@ntu.edu.sg.}
		
		\thanks{Manuscript received April 19, 2021; revised August 16, 2021.}}
	
	\markboth{Journal of \LaTeX\ Class Files,~Vol.~14, No.~8, August~2021}%
	{Shell \MakeLowercase{\textit{et al.}}: A Sample Article Using IEEEtran.cls for IEEE Journals}
	
	\maketitle
	
\begin{abstract}
Facial Action Unit (AU) detection seeks to recognize subtle facial muscle activations as defined by the Facial Action Coding System (FACS). A primary challenge w.r.t AU detection is the effective learning of discriminative and generalizable AU representations under conditions of limited annotated data. To address this, we propose a Hierarchical Vision-language Interaction for AU Understanding (HiVA) method, which leverages textual AU descriptions as semantic priors to guide and enhance AU detection. Specifically, HiVA employs a large language model to generate diverse and contextually rich AU descriptions to strengthen language-based representation learning. To capture both fine-grained and holistic vision-language associations, HiVA introduces an AU-aware dynamic graph module that facilitates the learning of AU-specific visual representations. These features are further integrated within a hierarchical cross-modal attention architecture comprising two complementary mechanisms: Disentangled Dual Cross-Attention (DDCA), which establishes fine-grained, AU-specific interactions between visual and textual features, and Contextual Dual Cross-Attention (CDCA), which models global inter-AU dependencies. This collaborative, cross-modal learning paradigm enables HiVA to leverage multi-grained vision-based AU features in conjunction with refined language-based AU details, culminating in robust and semantically enriched AU detection capabilities. Extensive experiments show that HiVA consistently surpasses state-of-the-art approaches. Besides, qualitative analyses reveal that HiVA produces semantically meaningful activation patterns, highlighting its efficacy in learning robust and interpretable cross-modal correspondences for comprehensive facial behavior analysis.
\end{abstract}
	
	\begin{IEEEkeywords}
		Facial Action Unit Detection, Vision Language Learning, Affective Computing
	\end{IEEEkeywords}
	
\section{Introduction}
\label{sec:intro}
	
		\begin{figure}[h]
		\centering
		\includegraphics[width=0.48\textwidth]{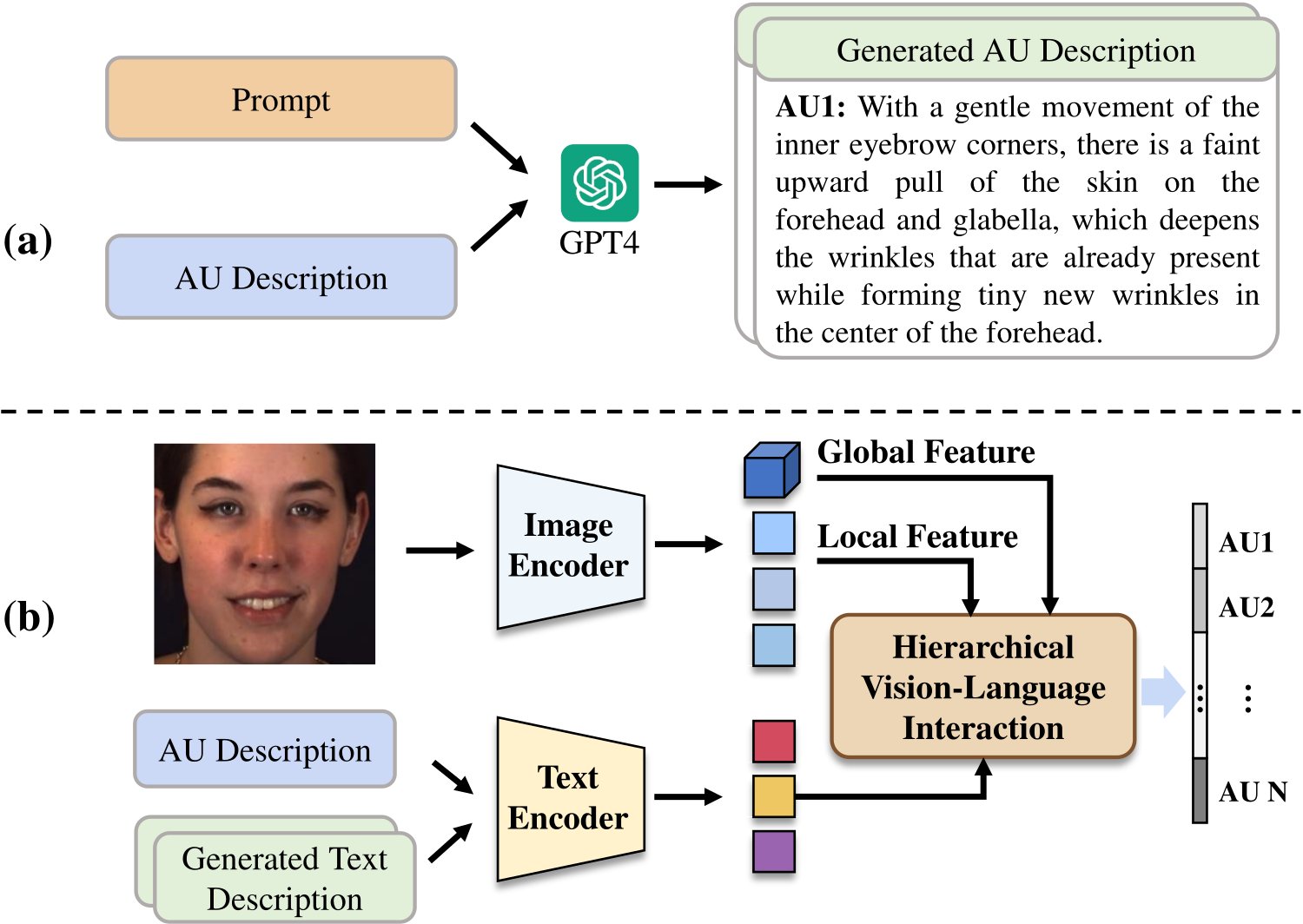}
		\caption{Overview of the proposed Hierarchical Vision-Language Interaction for Facial AU detection. (a) Illustrates the process of leveraging a large language model (e.g., GPT-4) to generate diverse and semantically rich AU descriptions, addressing the limitations of scarce textual knowledge. (b) depicts the core framework, which integrates visual features from input images and textual features from AU descriptions through hierarchical vision-language interaction, utilizing both local and global attention mechanisms for robust AU detection.
		}
		\label{fig:figure1}
	\end{figure}

Facial expressions are a vital component of nonverbal communication, providing rich insights into a person’s emotional and cognitive states~\cite{picard2000affective}. As a systematic method for describing and interpreting facial movements, the Facial Action Coding System (FACS)~\cite{friesen1978facial} has become the gold standard in both psychological research and computer vision. By decomposing complex expressions into individual Action Units (AUs), each corresponding to the activation of specific facial muscles, FACS enables a fine-grained and interpretable analysis of facial behavior~\cite{cohn2007observer}. This granular representation facilitates a deeper understanding of human affect and supports a wide range of applications, from emotion recognition and human-computer interaction to mental health monitoring and pain assessment~\cite{li2018occlusion, aboussalim2024pain, hamm2011automated}.
	
In recent years, the task of AU detection has witnessed significant advancements, largely driven by the rapid progress in deep learning techniques~\cite{li2017eac, li2017action, corneanu2018deep, shao2018deep, li2019semantic, yang2021exploiting, yang2023toward}. To learn AU-specific representations and improve the detection performance, many existing approaches utilize facial landmarks to localize or extract AU-relevant facial regions. Despite these improvements, deep learning-based supervised methods remain heavily reliant on large-scale annotated datasets. Models trained solely on visual data often suffer from overfitting to person-specific features, leading to limited generalization capabilities. To address the high cost and scarcity of fine-grained AU annotations, recent studies have explored semi-supervised and self-supervised learning paradigms~\cite{li2020learning, chang2022knowledge}, aiming to learn more generalizable AU representations. Nonetheless, most existing AU detection methods are purely vision-based, overlooking the potential of leveraging textual descriptions of AUs as a source of rich, complementary prior knowledge.

Textual descriptions of AUs are capable of enhancing AU detection robustness by providing structured prior knowledge that surpasses simple binary labels~\cite{yang2021exploiting, ge2024towards}. First, semantic priors improve AU localization and interpretation by detailing relevant facial regions (e.g., “chin boss,” “lip corner”), action types (e.g., “raise,” “press,” “pull”), motion attributes, and intra-/inter-AU relationships (e.g., mutual exclusivity of AU12 and AU15). These cues guide the model’s attention toward semantically informative areas, enhancing spatial focus while suppressing irrelevant activations. Second, textual descriptions facilitate more accurate modeling of inter-AU dependencies, capturing intra-AU semantics and inter-AU relations such as co-occurrence and exclusivity without manual dependency graphs. Thus, incorporating semantic descriptions allows the model to develop robust, context-aware AU representations. Furthermore, this paradigm would transform AU detection into a semantically grounded reasoning task, yielding more interpretable and resilient facial behavior analysis.

Building on these insights, we introduce a novel framework: Hierarchical Vision-language Interaction for AU Understanding (HiVA), aiming to overcome the limitations of purely vision-based AU detection methods by incorporating language-guided semantic grounding. As depicted in Fig.~\ref{fig:figure1}, HiVA adopts a unified architecture that integrates hierarchical cross-modal attention and interaction mechanisms to enhance both local and global AU understanding. The framework disentangles AU-specific representations and employs both fine-grained and coarse-grained attention to capture subtle relationships between visual features and AU-specific textual descriptions. 

To establish effective semantic cross-modal interaction between images and AU descriptions, we hypothesize that the model must first identify salient words within the descriptions that correspond to specific facial regions and muscle states, and subsequently construct fine-grained vision-language interactions. Accordingly, HiVA explicitly disentangles the representation learning of each AU under supervision from its corresponding label, thereby facilitating the extraction of discriminative AU-specific features and enabling precise one-to-one alignment between visual regions and textual semantics.

In addition to local cross-modal interaction, HiVA captures inter-AU contextual dependencies at the global facial level by incorporating the entire facial region into a global cross-modal interaction mechanism that aligns the global visual representation with all AU descriptions. This design enables the model to implicitly encode inter-AU relationships, allowing the prediction of one AU to be informed by both its localized region and other facial areas associated with semantically or functionally related AUs. 

Given that each AU is typically defined by only a few fixed sentences derived from FACS, we employ a large language model (LLM) to generate additional semantically consistent variants for each AU description. To mitigate potential semantic biases, all synthesized sentences are rigorously manually verified to ensure semantic accuracy, contextual appropriateness, and linguistic quality. Furthermore, we incorporate a regularization loss as a fault-tolerance mechanism to promote discriminative representations for each AU. This augmentation strategy expands the diversity of textual inputs, allowing the model to learn more robust and generalized AU-specific language representations. Consequently, the effectiveness of the hierarchical cross-modal attention is further enhanced, enabling more precise alignment between visual features and language-based AU descriptions.

Our main contributions can be summarized as follows:
\begin{itemize}
	\item We propose a unified framework that integrates hierarchical cross-modal attention and alignment to advance AU detection. HiVA jointly models vision and language modalities by grounding visual features in semantically rich AU descriptions.
	\item To establish effective and faithful vision-language correspondence, HiVA introduces dual attention mechanisms: fine-grained attention, which aligns individual AU descriptions with their corresponding visual regions; and coarse-grained attention, which captures inter-AU dependencies by associating global facial features with all AU descriptions. This dual-attention strategy jointly models both AU-specific semantics and inter-AU relational structures.
	\item Extensive experiments on multiple AU benchmarks demonstrate that HiVA consistently improves facial AU detection performance. Visualization analyses further confirm the emergence of semantically meaningful activation patterns for both individual and co-occurring AUs, underscoring HiVA’s effectiveness in capturing cross-modal correspondences.
\end{itemize}

The remainder of this paper is organized as follows. In Section \ref{sec:related work} we review related work in AU detection, vision and language learning. In Section \ref{sec:method}, we present the details of our HiVA method, while in Section \ref{sec:experiments}, we demonstrate its efficacy through experiments.

\section{Related Work}
\label{sec:related work}
\textbf{Facial action unit detection.} Automatic AU detection has been extensively studied for decades, leading to the development of numerous approaches~\cite{li2017action, li2017eac, corneanu2018deep, zhang2018identity, tu2019idennet, shao2018deep, li2019semantic, yang2020set, li2025decoupled, li2021meta, jacob2021facial, li2020learning, chen2022causal, yang2023toward, yu2025towards}. Since AUs reflect subtle changes in facial muscles and skin textures, many methods exploit facial landmarks~\cite{li2017action, li2017eac, corneanu2018deep, ge2024mgrr} or utilize adaptive attention mechanisms~\cite{shao2018deep, shao2019facial} to extract region-specific features. Another direction focuses on global-refinement strategies, where transformer-based multi-branch architectures~\cite{jacob2021facial} refine shared global facial representations via independent convolutional branches, and probabilistic graph models~\cite{song2021uncertain} model inter-AU dependencies and uncertainty to mitigate data imbalance.
Despite these advances, AU detection still faces a fundamental challenge: the annotation bottleneck caused by the costly and labor-intensive nature of FACS coding, which requires certified experts. To address this, self-supervised~\cite{chang2022knowledge, li2020learning} and weakly-supervised approaches have been explored. For example, Chang et al.~\cite{chang2022knowledge} proposed a contrastive learning framework that exploits inter-region differences to extract AU-related features while maintaining intra-region discrimination. In parallel, various methods explicitly model inter-AU relationships to improve detection performance, such as ME-GraphAU~\cite{luo2022learning}, which learns a dedicated AU graph capturing pairwise relationships among AUs.

Distinct from these vision-based approaches, the proposed HiVA framework introduces textual AU descriptions as semantic priors to guide detection. These linguistic cues focus attention on semantically relevant facial regions, suppress irrelevant activations, and enable semantically grounded reasoning, thereby enhancing both the interpretability and robustness of AU detection.

\textbf{Vision and Language Learning.} Recent multimodal pretraining work has shown that aligning images and text yields powerful, generalizable models, e.g., CLIP~\cite{radford2021learning} learns image features by predicting which caption matches which image over 400M web pairs, enabling flexible zero-shot transfer. BLIP~\cite{li2022blip} builds on this by bootstrapping noisy web captions (using a captioner and filter) to pretrain a model that excels at both vision-language understanding and generation. Flamingo ~\cite{alayrac2022flamingo} further combines frozen vision and language encoders with novel cross-attention layers, allowing in-context few-shot multimodal learning from arbitrarily interleaved image–text inputs. These advances have since been applied to facial behavior analysis, e.g., CLIPER ~\cite{li2024cliper} and DFER-CLIP~\cite{zhao2023prompting} adapt CLIP for static/dynamic facial expression recognition by using fine-grained textual expression descriptions as prompts.

Recently, several language-assisted AU detection methods have emerged. For instance, SEV-Net~\cite{yang2021exploiting} integrates visual features with AU semantic embeddings via cross-modal attention to improve AU detection performance. Subsequently, VL-FAU~\cite{ge2024towards} introduced a vision-language joint learning framework for explainable facial AU recognition, aiming to enhance both AU representation capability and language interpretability through the integration of multimodal tasks. In contrast to these existing vision-language AU detection approaches, the proposed HiVA framework incorporates hierarchical vision-language interaction, enabling it to capture subtle and complex associations between vision and language modalities. Furthermore, HiVA introduces dedicated architectural components for learning refined vision and language representations, such as disentangled AU-specific features (Sec.~\ref{sec:au_aware_dynamic}) and the regularization term designed to enhance the distinctiveness of inter-AU textual representations (Sec.~\ref{sec:hybrid_cnn_trans}).

\begin{figure*}[htb]
	\centering
	\includegraphics[width=0.9\textwidth]{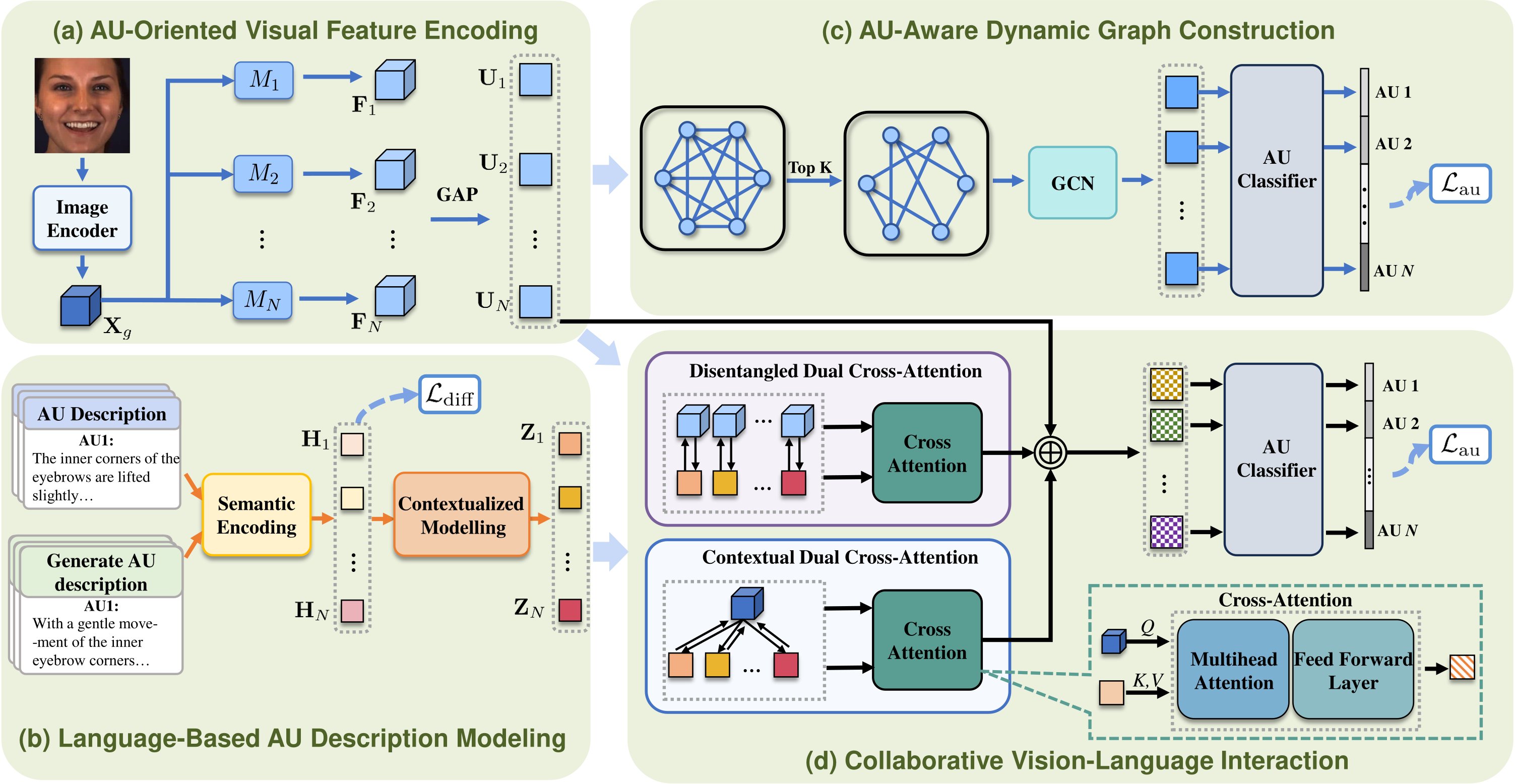}
	\caption{Framework of the proposed Hierarchical Vision-Language Attention for AU Understanding (HiVA). HiVA consists of four main components: (a) AU-Oriented Visual Feature Encoding, which extracts AU-specific visual representations using a CNN and transformer backbone; (b) Language-Based AU Description Modeling, which encodes textual AU descriptions to capture both intra- and inter-AU semantics; (c) AU-Aware Dynamic Graph Construction, which builds AU-specific graphs based on visual similarity to model adaptive inter-AU relations; and (d) Hierarchical Vision-Language Interaction, which employs Disentangled Dual Cross-Attention (DDCA) and Contextual Dual Cross-Attention (CDCA) to interact visual and textual features at both local and global levels.
	}
	\label{fig:framework}
\end{figure*}

\section{Method}
\label{sec:method}


\subsection{Framework Overview}
We propose a Hierarchical Vision-Language Attention for AU Understanding method, i.e., \texttt{HiVA}, to incorporate textual AU descriptions into AU detection, aiming to model the intra- and inter-AU relationships and perceive co-occurrence and exclusivity patterns without the need for manually designed dependency structures. The architecture of the HiVA is shown in Fig.~\ref{fig:framework}. 

Specifically, HiVA is organized into four sequential modules. (1) \textbf{AU-Oriented Visual Feature Encoding} extracts hierarchical visual representations from the input facial images. (2) \textbf{Language-Based AU Description Modelling} encodes the accompanying textual AU descriptions at both token and sentence levels, thus preserving fine-grained semantic cues and capturing the intricate intra- and inter-AU dependencies present in the linguistic modality. (3) \textbf{AU-Aware Dynamic Graph Construction} utilizes the vision-derived AU representations to build sample-specific graphs, which promote the learning of discriminative AU-specific features and provide the foundation for disentangled dual cross-attention between visual and textual streams. (4) \textbf{Hierarchical Vision-Language Interaction} fuses global and local AU features with their contextualised language counterparts through complementary cross-attention mechanisms, jointly leveraging multi-granular visual cues and semantically enriched textual information. Together, these components furnish a coherent and robust framework for semantically informed facial AU detection.

Below, we present details of the four parts of HiVA. 

\subsection{Image and Textual AU Feature Encoding}
\label{sec:hybrid_cnn_trans}
The proposed HiVA framework simultaneously leverages facial images and textual descriptions of AUs to jointly learn vision- and language-based AU representations. As illustrated in Fig.~\ref{fig:framework}, facial images are processed through a convolutional neural network backbone to extract visual feature maps capable of capturing the subtle and dynamic activation or inactivation states of individual AUs. To obtain discriminative, region-specific representations for each AU, we employ independent neural branches per AU and incorporate a graph neural network to explicitly model inter-AU relationships within the visual modality, thus enhancing the discriminative power of the AU-specific features.

For the encoding of textual AU representations, the framework integrates two key components: a Semantic Encoding module and a Contextualized Modeling module. The Semantic Encoding module aggregates token-level features within each AU description to capture intra-AU semantics. In contrast, the Contextualized Modeling module is designed to capture inter-AU relationships in the language domain, enabling the learning of contextualized textual AU features. These features are subsequently employed in the vision-language interaction module (Sec.~\ref{sec:vision_language_interaction}). The following sections provide a detailed exposition of each component.


\textbf{Image feature encoding}. We employ the Swin Transformer as the visual backbone for image feature extraction, leveraging its hierarchical architecture and shifted window-based self-attention mechanism to effectively capture local and global contextual information. The model processes the input image through four successive stages, incorporating patch merging operations between stages to progressively reduce spatial resolution while expanding the feature dimensionality.

Given an input facial image $\mathcal{I} \in \mathbb{R}^{ 224 \times 224 \times 3}$, the image is first partitioned into non-overlapping patches of size $4 \times 4$, yielding  $56 \times 56$ patches in total. Each patch is flattened and linearly projected into a 128-dimensional embedding, resulting in an initial feature representation of shape $56 \times 56 \times 128$.
Subsequently, the hierarchical encoder produces multi-scale feature maps $\mathbf{X}$ for stages ${i\in \{1,2,3,4\}}$, with spatial and channel dimensions of $56 \times 56 \times 128$, $28 \times 28 \times 256$,  $14 \times 14 \times 512$, and $7 \times 7 \times 1024$, respectively. To align the visual feature representation with the textual modality, the output from the final stage is further processed using a pointwise  convolution, yielding a transformed feature map $\mathbf{X}_g \in \mathbb{R}^{ 7 \times 7 \times 512}$. This transformation ensures dimensional compatibility between the visual and textual embeddings, enabling the subsequent cross-modal alignment operations, which will be detailed in the following sections.

\textbf{Textual Encoding of AU Descriptions}. 
To encode prior AU knowledge embedded in the textual AU description, we incorporate Semantic Encoding of AU descriptions to capture semantic information from individual AU textual descriptions using a language model like BERT, which is followed by Contextualized Modeling of inter-AU Dependencies aimed to learn how different AUs interact or relate within a shared representational space.

\textit{Semantic Encoding of AU Description}.
Let $\mathcal{T} = \{T_1, T_2, \dots, T_N\}$ denote the set of textual descriptions corresponding to $N$ predefined AUs, where each $T_i$ is a natural language description for AU$_i$. 
Specifically, each AU description is first tokenized using the WordPiece tokenizer, and the resulting tokens are embedded via a combination of trainable word embeddings, segment embeddings, and positional encodings, generating a token sequence for each AU description $\widehat{\mathcal{T}} = \{\widehat{T}_1, \widehat{T}_2, \dots, \widehat{T}_N\}$. 
At this stage, each $\widehat{T}_i$ is encoded using a pre-trained BERT-large model, denoted as $\mathrm{BERT}_{\mathrm{large}}(\cdot)$, to capture contextualized intra-AU semantic representations. This process can be formulated as $H_i = \mathrm{BERT}_{\mathrm{large}}(\widehat{T}_i) \in \mathbb{R}^{L_i \times d'}$, where $L_i$ is the number of tokens in $\widehat{T}_i$, and $d' = 1024$ is the hidden dimension of BERT-large. The final representation for each AU is obtained by averaging the contextualized token embeddings from the last Transformer layer, yielding a fixed-length sentence-level semantic vector $\mathbf{H}_i = \frac{1}{L_i} \sum_{j=1}^{L_i} H_i^{(j)} \in \mathbb{R}^{d'}$. The resulting intra-AU representations for all AUs are aggregated as $\mathcal{H} = [\mathbf{H}_1, \mathbf{H}_2, \dots, \mathbf{H}_N]^\top \in \mathbb{R}^{N \times d'}$.

Considering the scarcity of AU descriptions, we incorporate a regularization loss here to explicitly ensure that the textual representations w.r.t different AU should be distinguishable:

\begin{equation}
	\mathcal{L}_{\text{diff}} = \frac{1}{N^2} \left\| \left( \frac{\mathcal{H}}{\|\mathcal{H}\|_2} \right) \left( \frac{\mathcal{H}}{\|\mathcal{H}\|_2} \right)^\top - E \right\|_F^2,
	\label{equ:intra_au_diff}
\end{equation}
where $\frac{\mathcal{H}}{\|\mathcal{H}\|_2}$ denotes row-wise $\ell_2$ normalization of $\mathcal{H}$. $E \in \mathbb{R}^{N \times N}$ is the identity matrix. $\|\|_F^2$ denotes the Frobenius norm.

\textit{Contextualized Modeling of inter-AU Dependencies}.
To capture semantic dependencies and relational structures among AUs, we introduce a three-layer Transformer encoder, denoted as $\mathrm{Trans}(\cdot)$, which operates on the set of AU-level embeddings $\tilde{H}$. Each Transformer layer comprises multi-head self-attention and position-wise feedforward sublayers, along with residual connections and layer normalization.
	
The Contextualized Modeling of inter-AU Dependencies operation can be formalized as:
\begin{equation}
	\mathcal{Z} = \mathrm{Trans}(\mathcal{H}) \in \mathbb{R}^{N \times d},
\end{equation}
where $d = 512$ is the target embedding dimension. The output $\mathcal{Z} = [\mathbf{Z}_1, \mathbf{Z}_2, \dots, \mathbf{Z}_N]^\top$ consists of enriched representations for all AU descriptions, capturing both the intra-AU semantics and inter-AU contextual dependencies.

\subsection{AU-Aware Dynamic Graph Construction}
\label{sec:au_aware_dynamic}

To learn AU-specific representations, a set of $N$ independent processing branches is employed. Each branch consists of a pointwise convolution layer followed by global average pooling (GAP). The pointwise convolution layer transforms the shared representation $\mathbf{X}_g$ into an AU-specific feature map $\mathbf{F}_i \in \mathbb{R}^{H \times W \times C}$, which is then reduced via GAP to a compact vector $\mathbf{U}_i \in \mathbb{R}^C$. This process yields a set of AU-specific features $\mathcal{U} = \{\mathbf{U}_1, \mathbf{U}_2, \ldots, \mathbf{U}_N\}$, each encoding both local activation cues and the contextual facial information relevant to the corresponding AU.

To capture the relational structure among AUs, we compute pairwise similarities between node features. For each AU $i$, the similarity to another AU $j$ is computed via inner product $\text{Sim}(i, j) = \mathbf{U}_i^\top \mathbf{U}_j$.
Subsequently, a directed graph is constructed by connecting each node to its top $K$ most similar neighbors. Once the graph is established, a graph convolutional operation is performed to refine the node features. The updated representation $\mathbf{v}_i^{\text{upd}}$ for AU $i$ is computed as:
\begin{equation}
	\mathbf{U}_i^{\text{upd}} = \sigma \left[ \mathbf{U}_i + g\left(\mathbf{U}_i, \sum_{j=1}^{N} r(\mathbf{U}_j, a_{i,j}) \right) \right],
\end{equation}
where $\sigma(\cdot)$ denotes a non-linear activation function, $g(\cdot)$ and $r(\cdot, \cdot)$ are learnable transformations associated with the graph convolution, and $a_{i,j} \in \{0,1\}$ indicates whether AU$_j$ is connected to AU$_i$. Incorporating graph construction into HiVA enables modeling of dynamic, context-dependent relationships among AUs, which in turn enhances AU-specific feature learning and improves detection accuracy.

To predict the activation of each AU, we apply a fully connected (FC) layer directly to the updated node representation of each AU to obtain a scalar prediction score $p_i$. We exploit the multi-label sigmoid cross-entropy loss for AU detection. It is formulated as
\begin{equation}
	\mathcal{L}_{\text{au}} = - \sum^{N}_{i} y_i \log p_i + (1 - y_i) \log (1- p_i)
	\label{equ:au_loss}
\end{equation}
where $N$ is the number of AUs. $y_i$ denotes the $i$-th ground truth AU label of the input AU sample. 

\subsection{Hierarchical Vision-Language Interaction}
\label{sec:vision_language_interaction}
\begin{figure}[h]
	\centering
	\includegraphics[width=0.48\textwidth]{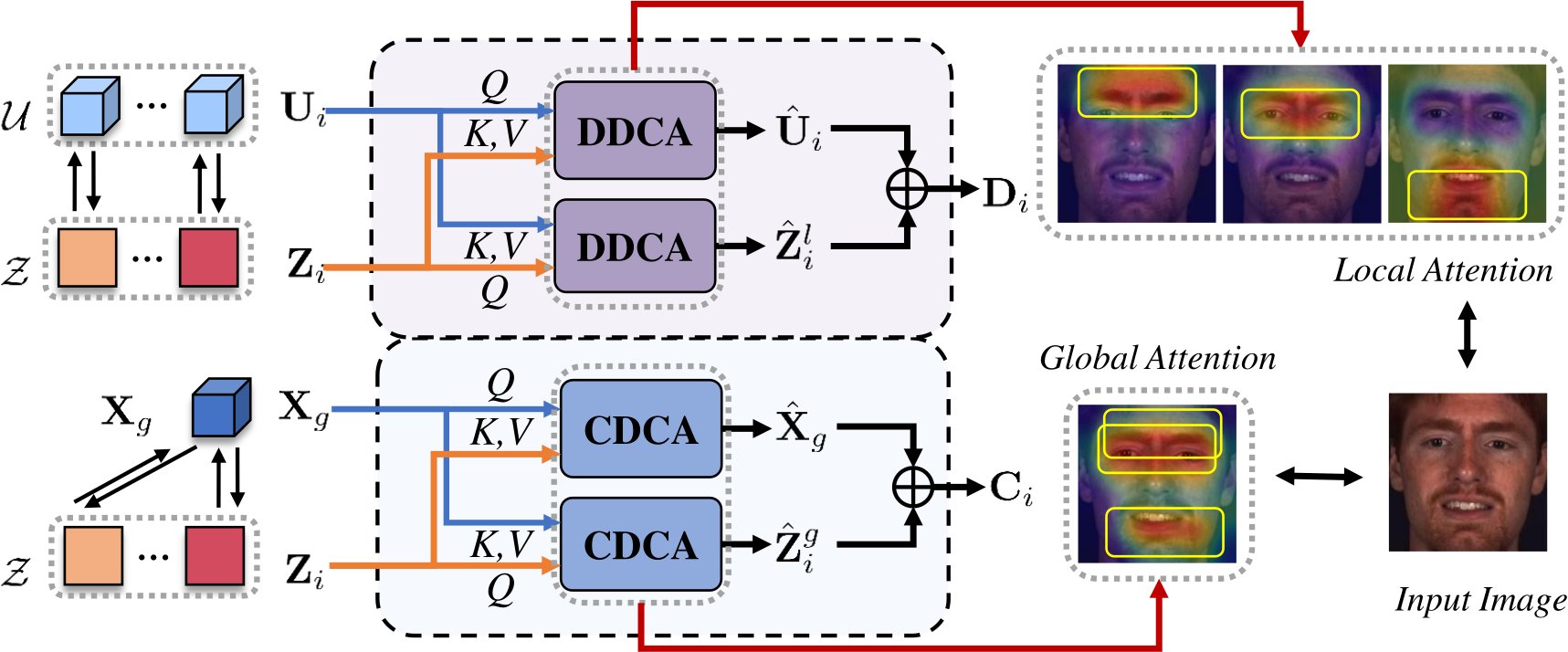}
	\caption{ Illustration of the Hierarchical Vision-Language Interaction module in HiVA. This component integrates two complementary attention mechanisms: Disentangled Dual Cross-Attention (DDCA) for fine-grained one-to-one interaction between AU-specific visual features and corresponding textual descriptions, and Contextual Dual Cross-Attention (CDCA) between global visual features and all AU descriptions. This bidirectional, cross-modal interaction enables the model to capture both localized semantic grounding and global inter-AU dependencies for robust AU detection.
	}
	\label{fig:figure3}
\end{figure}

In this stage, HiVA takes the global AU feature $\mathbf{X}_g$ as well as the disentangled AU-specific representations $\mathcal{U}$ to perform cross-attention with the contextual language-based AU features $\mathcal{Z}$. Typically, Disentangled Dual Cross-Attention (DDCA) is performed among $\mathbf{U}_i$ and $\mathcal{Z}$ to  establish semantic one-to-one cross-modal interaction between images and AU descriptions. Similarly, Contextual Dual Cross-Attention (CDCA) will be conducted among $\mathbf{X}_g$ and $\mathcal{Z}$ to reason about inter-AU relationships and spatial overlap across facial regions. 
These two mechanisms form a bidirectional and hierarchical vision-language learning framework, allowing the model to leverage both global and local information in a cross-modal, context-driven manner. We illustrate the pipeline of Hierarchical Vision-Language Interaction in Fig.~\ref{fig:figure3}.

\subsubsection{DDCA}
To establish a precise and semantically faithful cross-modal interaction between facial images and AU descriptions, it is essential that the model accurately determines the activation state of each AU based on salient textual cues that correspond to specific facial regions and muscle configurations. To this end, HiVA disentangles the representation learning process for each AU and derives AU-specific latent features, as illustrated in Fig.~\ref{fig:figure3}. This design enables explicit one-to-one alignment between localized visual features and their corresponding textual descriptions.

Leveraging the encoded AU-specific visual and textual representations, denoted as $\mathbf{U}_i$ and $\mathbf{Z}_i$ respectively, HiVA adopts a transformer-based one-to-one cross-attention mechanism, where the visual feature $\mathbf{U}_i$ serves as the query and the corresponding textual feature $\mathbf{Z}_i$ is used to generate both the key and value. The attention output is computed as:

\begin{equation}
	\hat{\mathbf{U}}_i = \text{Softmax}\left( \frac{(\mathbf{W}_Q^L \mathbf{U}_i)^\top (\mathbf{W}_K^L \mathbf{Z}_i)}{\sqrt{d}} \right) \cdot (\mathbf{W}_V^L \mathbf{Z}_i),
\end{equation}
where $\mathbf{W}_Q^L, \mathbf{W}_K^L, \mathbf{W}_V^L \in \mathbb{R}^{d \times C}$ are learnable projection matrices, and $d$ denotes the dimensionality of the attention space. This vision-language cross-attention mechanism enables the model to selectively focus on visual regions that are semantically aligned with the AU description (e.g., muscle group, movement direction, or facial location), thus improving localization accuracy, enhancing feature discriminability, and promoting semantic interpretability. 

As illustrated in Fig.~\ref{fig:figure3}, DDCA further introduces a dual cross-attention module that reverses the roles of vision and language: the textual representation $\mathbf{Z}_i$ serves as the query, while the visual feature $\mathbf{U}_i$ is used as the key and value, yielding refined textual features $\hat{\mathbf{Z}}^l_i$. This complementary attention mechanism allows the model to adaptively refine language representations conditioned on the visual context.  Finally, the refined visual feature $\hat{\mathbf{U}}_i$ is pooled into a one-dimensional vector and combined with $\hat{\mathbf{Z}}^l_i$ via element-wise addition to produce the final representation $\mathbf{D}_i$ for the $i$-th AU.

\subsubsection{CDCA}
To capture contextual dependencies between global facial features and all AU textual descriptions, we introduce a Contextual Dual Cross-Attention (CDCA) mechanism. Let $\mathbf{X}_g \in \mathbb{R}^{H \times W \times C}$ denote the global visual feature map extracted from the entire facial image, where $H$ and $W$ represent the spatial dimensions and $C$ is the number of feature channels. The textual embedding of the $i$-th AU is denoted by $\mathbf{Z}_i \in \mathbb{R}^{C}$.

In the first stage of CDCA, for each AU $i$, the textual feature $\mathbf{Z}_i$ is employed as the query, while the global visual feature map $\mathbf{X}_g$ is flattened into a sequence of spatial tokens, denoted as $\mathbf{X}_g^{\text{flat}} \in \mathbb{R}^{(H \cdot W) \times C}$, and used to compute keys and values. The cross-modal attention output is formulated as:
\begin{equation}
\hat{\mathbf{Z}}_i = \text{Softmax} \left( \frac{(\mathbf{W}_Q^G \mathbf{Z}_i)^\top (\mathbf{W}_K^G \mathbf{X}_g^{\text{flat}})}{\sqrt{d}} \right) \cdot (\mathbf{W}_V^G \mathbf{X}_g^{\text{flat}}),
\end{equation}
where $\mathbf{X}_g^{\text{flat}} \in \mathbb{R}^{(H \cdot W) \times C}$ is the flattened visual feature map, and $\mathbf{W}_Q^G, \mathbf{W}_K^G, \mathbf{W}_V^G \in \mathbb{R}^{d \times C}$ are learnable projection matrices. The resulting context-aware representation $\hat{\mathbf{Z}}_i$ integrates semantic information from all relevant spatial locations, thereby enriching the representation of AU $i$ with global visual context. This attention  formulation facilitates the modeling of inter-AU relationships and spatial overlaps across facial regions, contributing to more robust and accurate AU detection.

Similarly, to further enhance context-aware AU representation learning, CDCA reverses the attention flow: the global visual tokens $\mathbf{X}_g^{\text{flat}}$ serve as queries, while the full set of AU textual embeddings $\mathcal{Z}$ are employed as keys and values to produce refined visual features $\hat{\mathbf{X}}_g$. The refined features $\hat{\mathbf{X}}_g$ are then pooled into one-dimensional vectors and element-wise combined with $\hat{\mathbf{Z}}^g_i$ to obtain the generated features $\mathbf{C}_i$ for the $i$-th AU. By integrating global visual features with all AU-specific textual descriptions via dual cross-attention, CDCA explicitly models inter-AU dependencies, enabling each AU description to attend to spatial regions that may be shared or jointly activated across multiple AUs.

As illustrated in Fig.~\ref{fig:framework}, the final AU representation is obtained by concatenating $\mathbf{U}_i$, $\mathbf{D}_i$, and $\mathbf{C}_i$, which is subsequently projected into a 512-dimensional latent space through a fully connected layer for AU prediction.


\subsection{Training Strategy.} 
We adopt a two-stage training strategy to progressively optimize the HiVA framework.

In the first stage, we jointly train the AU-Oriented Visual Feature Encoding and AU-Aware Dynamic Graph Construction modules, i.e., (a) and (c) in Fig.~\ref{fig:framework}. The objective of this stage is to learn robust global AU representations, along with AU-specific features that encode both activation states and inter-AU associations for each facial expression, establishing the foundation for subsequent AU-oriented vision-language interactions. At this stage, the training objective is $\mathcal{L}_{\text{au}}$ and has been formulated in Eq.~\ref{equ:au_loss}.

In the second stage, we train the complete HiVA architecture by integrating all three components: AU-Oriented Visual Feature Encoding, Language-Based AU Description Modeling, and Hierarchical Vision-Language Interaction, i.e., (a), (b) and (d) in Fig.~\ref{fig:framework}. By leveraging fine-grained and global cross-attention mechanisms, this stage aligns visual features with semantic descriptions to capture subtle AU relationships, significantly improving detection accuracy and robustness. At this stage, the training objective is the linear combination of $\mathcal{L}_{\text{au}}$ and $\mathcal{L}_{\text{diff}}$:
\begin{equation}
	\mathcal{L}_{\text{tot}} = \mathcal{L}_{\text{au}} + \lambda \mathcal{L}_{\text{diff}},
\end{equation}
where $\lambda$ means the trade-off factor that control the importance of the
regularization loss formulated in Eq.~\ref{equ:intra_au_diff}.
To address the prevalent class imbalance prevalent in AU datasets, we apply inverse frequency-based reweighting to the loss function. Specifically, samples from under-represented AUs are assigned higher weights, thus penalizing the misclassification of rare classes more heavily and encouraging the model to learn discriminative features across all AU categories. The overall training process is illustrated in Alg.~\ref{alg:hiva}. Code and the pretrained model will be publicly available on GitHub.

\begin{algorithm}[htb]
	\small
	\SetAlgoLined
	\KwIn{AU images $\mathcal{I}$, textual AU descriptions $\mathcal{T}$, ground truth AU labels $y$}
	\KwOut{Optimized model parameters $\theta$}
	// 	Stage 1: Global and Local AU feature Encoding \\
	\While{not converge}{
		
		Extract hierarchical vision features $\mathbf{X}_g$ from  $\mathcal{I}$ \\
		
		Extract AU-specific features $\mathcal{U} = \{\mathbf{U}_1, \mathbf{U}_2, \ldots, \mathbf{U}_N\}$ from $\mathbf{X}_g$ \\
		Compute graph refined AU features via Eq.~3 \\
		
		Calculate AU detection loss $\mathcal{L}_{\text{au}}$ via Eq.~5 \\
		Optimize part of $\theta$ w.r.t Fig. 2 (a) and (c) with $\mathcal{L}_{\text{au}}$ \\
	}
	
	// 	Stage 2: Hierarchical Vision-Language Interaction \\
	\While{not converge}{
		Generate intra-AU representations $\mathcal{H}$ from $\mathcal{T}$ \\
		Generate contextualized textual AU representations $\mathcal{Z}$ from $\mathcal{H}$ \\
		Extract hierarchical vision features $\mathbf{X}_g$ from  $\mathcal{I}$ \\
		
		Extract AU-specific features $\mathcal{U}$ from $\mathbf{X}_g$\\
		
		Generate DDCA-enhanced local AU features $\mathbf{D}_i$ from $\mathcal{U}$ and $\mathcal{Z}$ \\
		Generate CDCA-enhanced global AU features $\mathbf{C}_i$ from $\mathbf{X}_g$ and  $\mathcal{Z}$ \\
		Calculate  $\mathcal{L}_{\text{au}}$ and $\mathcal{L}_{\text{diff}}$ \\
		Update total loss $\mathcal{L} \leftarrow \mathcal{L}_{\text{au}} + \lambda \mathcal{L}_{\text{diff}}$ \\
		Optimize $\theta$ using total loss $\mathcal{L}$ \\
	}
	\caption{Training Pipeline for HiVA}
	\label{alg:hiva}
\end{algorithm}

\begin{table*}[htb]
	\caption{F1 scores on BP4D dataset.  \textbf{Bold} and \underline{underline} denote the best and second best results, respectively.}
	\centering
	\label{tab:bp4d_cross_dataset}
	\begin{tabular}{c|c|c|c|c|c|c|c|c|c|c|c|c|c}
	\toprule
		\textbf{Methods/AU} & 1 & 2 & 4 & 6 & 7 & 10 & 12 & 14 & 15 & 17 & 23 & 24 & \textbf{AVE} \\
		\midrule
		EAC-Net\cite{li2017eac} &  39.0 & 35.2 & 48.6 & 76.1 & 72.9 & 81.9 & 86.2 & 58.8 & 37.5 & 59.1 & 35.9 & 35.8 & 55.9 \\
		ATF \cite{zhang2018identity} &  39.2 & 35.2 & 45.9 & 71.6 & 71.9 & 79.0 & 83.7 & 65.5 & 33.8 & 60.0 & 37.3 & 41.8 & 55.4 \\
		IdenNet \cite{tu2019idennet} &  50.5 & 35.9 & 50.6 & 77.2 & 74.2 & 82.9 & 85.1 & 65.3 & 42.2 & 60.8 & 42.1 & 46.5 & 59.3 \\
		DSIN \cite{corneanu2018deep} &  51.7 & 40.4 & 56.0 & 76.1 & 73.5 & 79.9 & 85.4 & 62.7 & 37.3 & 62.9 & 38.8 & 41.6 & 58.9 \\
		JAA-Net\cite{shao2018deep}  & 47.2 & 44.0 & 54.9 & 77.5 & 74.6 & 84.0 & 86.9 & 61.9 & 43.6 & 60.3 & 42.7 & 41.9 & 60.0 \\
		ARL \cite{shao2019facial} & 45.8 & 39.8 & 55.1 & 75.7 & 77.2 & 82.3 & 86.6 & 58.8 & 47.6 & 62.1 & 47.4 & 55.4 & 61.1 \\
		SRERL \cite{li2019semantic} & 46.9 & 45.3 & 55.6 & 77.1 & 78.4 & 83.5 & 87.6 & 63.9 & 52.2 & 63.9 & 47.1 & 53.3 & 62.1 \\
		UGN \cite{song2021uncertain} & 54.2 & 46.4 & 56.8 & 76.2 & 76.7 & 82.4 & 86.1 & 64.7 & 51.2 & 63.1 & 48.5 & 53.6 & 63.3 \\		
		Jacob \textit{et al.} \cite{jacob2021facial} & 51.7 & 49.3 & 61.0 & 77.8 & 79.5 & 82.9 & 86.3 & 67.6 & 51.9 & 63.0 & 43.7 & 56.3 & 64.2 \\
		SO-Net \cite{yang2020set} & 40.2 & 46.2 & 56.0 & 79.3 & 73.5 & 84.2 & \textbf{90.8} & 64.7 & \underline{55.9} & 61.0 & 37.4 & 40.2 & 60.8 \\
		TAE \cite{li2020learning} &  47.0 & 45.9 & 50.9 & 74.7 & 72.0 & 82.4 & 85.6 & 62.3 & 48.1 & 62.3 & 45.9 & 46.3 & 60.3 \\
		PIAP \cite{tang2021piap} & 54.2 & 47.1 & 54.0 & 79.0 & 78.2 & \underline{86.3} & \underline{89.5} & 66.1 & 49.7 & 63.2 & 49.4 & 52.0 & 64.1\\
		CISNet~\cite{chen2022causal} & 54.8 & 48.3 & 57.2 & 76.2 & 76.5 & 85.2 & 88.4 & 66.2 & 50.9 & \underline{65.0} & 47.7 & 56.5 & 64.3 \\
		AUNet~\cite{yang2023toward} & \underline{58.0} & 48.2 & 62.4 & 76.4 & 77.5 & 83.4 & 88.4 & 63.4 & 52.0 & \textbf{65.5} & 52.1 & 52.3 & 65.0 \\
		ME-GraphAU~\cite{luo2022learning} & 52.7 & 44.3 & 60.9 & \underline{79.9} & \textbf{80.1} & 85.3 & 89.2 & \textbf{69.4} & 55.4 & 64.4 & 49.8 & 55.1 & 65.5 \\
		AAR~\cite{shao2023facial} & 53.2 & 47.7 & 56.7 & 75.9 & 79.1 & 82.9 &88.6 & 60.5 & 51.5 & 61.9 & 51.0 & 56.8 & 63.8 \\
		MGRR-Net~\cite{ge2024mgrr} & 52.6 & 47.9 & 57.3 & 78.5 & 77.6 & 84.9 & 88.4 & 67.8 & 47.6 & 63.3 & 47.4 & 51.3 & 63.7 \\
		MDHR~\cite{wang2024multi} & 54.6 & 49.7 & 61.0 & 79.9 & 79.4 & 85.4 &88.5 & 67.8 & 56.8 & 63.2 & 50.9 & 55.4 & 66.1 \\
		\midrule
		SEV-Net~\cite{yang2021exploiting}  &  \textbf{58.2} & \textbf{50.4} & 58.3 & \textbf{81.9} & 73.9 & \textbf{87.8} & 87.5 & 61.6 & 52.6 & 62.2 & 44.6 & 47.6 & 63.9 \\
		CLEF~\cite{zhang2023weakly} & 55.8 & 46.8 & \textbf{63.3} & 79.5 & 77.6 & 83.6 & 87.8 & 67.3 & 55.2 & 63.5 & \underline{53.0} & \textbf{57.8} & \underline{65.9} \\
		VL-FAU~\cite{ge2024towards}  &  56.3 & \underline{49.9} & \underline{62.6} & 79.5 & \textbf{80.1} & 82.6 & 88.6 & 66.8 & 51.3 & 63.5 & 51.3 & \underline{57.1} & 65.8 \\
		\textbf{HiVA (Ours)}  &  54.3 & 49.7 & \textbf{63.3} & 79.3 & \underline{79.8} & 84.5 & 88.8 & \underline{68.5} & \textbf{57.0} & 62.6 & \textbf{53.1}  & 56.8 & \textbf{66.5} \\
		\bottomrule
	\end{tabular}
\end{table*}

\begin{table*}[htb]
	\centering
	\caption{F1 scores on DISFA dataset.  \textbf{Bold} and \underline{underline} denote the best and second best results, respectively.}
	\label{tab:disfa_cross_dataset}
	\begin{tabular}{c|c|c|c|c|c|c|c|c|c}
	\toprule
		\textbf{Methods/AU} & 1 & 2 & 4 & 6 & 9 & 12 & 25 & 26 & \textbf{AVE} \\
	\midrule
		EAC-Net \cite{li2017eac}  & 41.5 & 26.4 & 66.4 & 50.7 & \textbf{80.5} & \textbf{89.3} & 88.9 & 15.6 & 48.5 \\
		ATF \cite{zhang2018identity}  & 45.2 & 39.7 & 47.1 & 48.6 & 32.0 & 55.0 & 86.4 & 39.2 & 49.2 \\
		IdenNet \cite{tu2019idennet} & 25.5 & 34.8 & 64.5 & 45.2 & 44.6 & 70.7 & 81.0 & 55.0 & 52.6 \\ 
		DSIN \cite{corneanu2018deep}  & 42.4 & 39.0 & 68.4 & 28.6 & 46.8 & 70.8 &  90.4 & 42.2 & 53.6 \\
		JAA-Net \cite{shao2018deep}  & 43.7 & 46.2 & 56.0 & 41.4 & 44.7 & 69.6 & 88.3 & 58.4 & 56.0 \\
		ARL \cite{shao2019facial} & 43.9 & 42.1 & 63.6 & 41.8 & 40.0 & 76.2 & 95.2 & \underline{66.8} & 58.7 \\ 
		SRERL \cite{li2019semantic} & 45.7 & 47.8 & 59.6 & 47.1 & 45.6 & 73.5 & 84.3 & 43.6 & 55.9 \\
		UGN \cite{song2021uncertain} & 43.3 & 48.1 & 63.4 & 49.5 & 48.2 & 72.9 & 90.8 & 59.0 & 60.0 \\
		Jacob \textit{et al.} \cite{jacob2021facial} & 46.1 & 48.6 & 72.8 & \underline{56.7} & 50.0 & 72.1 & 90.8 & 55.4 & 61.5 \\
		SO-Net \cite{yang2020set} &  33.8 & 44.5 & 70.3 & \textbf{57.6} & 39.7 & 78.2 & 86.7 & 57.3 & 58.5 \\
		PIAP \cite{tang2021piap} & 50.2 & 51.8 & 71.9 & 50.6 & 54.5 & 79.7 & 94.1 & 57.2 & 63.8\\
		TAE \cite{li2020learning} &  21.4 & 19.6 & 64.5 & 46.8 & 44.0 & 73.2 & 85.1 & 55.3 & 51.5 \\
		CISNet~\cite{chen2022causal} & 48.8 & 50.4 & \textbf{78.9} & 51.9 & 47.1 & \underline{80.1} & \underline{95.4} & 65.0 & 64.7 \\
		AUNet~\cite{yang2023toward}  & 60.3 & 59.1 & 69.8 & 48.4 & 53.0 & 79.7 & 93.5 & 64.7 & 66.1 \\
		ME-GraphAU~\cite{luo2022learning} &  54.6 & 47.1 & 72.9  & 54.0 & 55.7 & 76.7 & 91.1 & 53.0 & 63.1 \\
		AAR~\cite{shao2023facial} & 62.4 & 53.6 & 71.5 & 39.0 & 48.8 & 76.1 & 91.3 & 70.6 & 64.2  \\
		MGRR-Net~\cite{ge2024mgrr} & \underline{61.3} & \textbf{62.9} & \underline{75.8} & 48.7 & 53.8 & 75.5 & 94.3 & \textbf{73.1} & \textbf{68.2} \\
		MDHR~\cite{wang2024multi} & 65.4 & 60.2 & 75.2 & 50.2 & 52.4 & 74.3 & 93.7 & 58.2 & 66.2 \\
	\midrule
	SEV-Net~\cite{yang2021exploiting}  &  55.3 & 53.1 & 61.5 & 53.6 & 38.2 & 71.6 & \textbf{95.7} & 41.5 & 58.8 \\
	CLEF~\cite{zhang2023weakly} & \textbf{64.3} & \underline{61.8} & 68.4 & 49.0 & 55.2 & 72.9 & 89.9 & 57.0 & 64.8 \\
	VL-FAU~\cite{ge2024towards} & 60.9 & 56.4 & 74.0 & 46.3 & 60.8 & 72.4 & 94.3 & 66.5 & 66.5 \\
	\textbf{HiVA (Ours)}  &  60.6 & 58.4 & 75.4 & 51.0 & \underline{61.2} & 74.8 & 93.9 & 63.8 & \underline{67.4} \\
		\bottomrule
	\vspace{-1.5em}
	\end{tabular}
\end{table*}

\begin{table*}[htb]
	\centering
	\caption{F1 score on GFT dataset. \textbf{Bold} and \underline{underline} denote the best and second best results, respectively.}
	\label{tab:gft_cross_dataset}
	\begin{tabular}{c|c|c|c|c|c|c|c|c|c|c|c}
		\toprule
		\textbf{Methods/AU} & 1 & 2 & 4 & 6 & 10 & 12 & 14 & 15 & 23 & 24 & \textbf{AVE} \\
		\midrule
		AlexNet \cite{girard2017sayette}  &  44 & 46 & 2 & 73 & 72 & 82 & 5 & 19 & 43 & 42 & 42.8  \\
		EAC-Net~\cite{li2017eac} & 15.5 & 56.6 & 0.1 & 81.0 & 76.1 & 84.0 & 0.1 & 38.5 & 57.8 & \underline{51.2} & 46.1 \\
		ARL~\cite{shao2019facial} & 51.9 & 45.9 & 13.7 & 79.2 & 75.5 & 82.8 & 0.1 & 44.9 & 59.2 & 47.5 & 50.1 \\
		Ertugrul et al.~\cite{ertugrul2020crossing} & 43.7 & 44.9 & 19.8 & 74.6 & 76.5 & 79.8 & \textbf{50.0} & 33.9 & 16.8 & 12.9 & 45.3 \\
		AAR~\cite{shao2023facial} & \textbf{66.3} & 53.9 & 23.7 & \underline{81.5} & 73.6 & 84.2 & \underline{43.8} & \textbf{53.8} & \underline{58.2} & 46.5 & 58.5 \\
		MAL \cite{li2021meta} & 52.4 & 57.0 & \textbf{54.1} & 74.5 & \underline{78.0} & 84.9 & 43.1 & 47.7 & 54.4 & \textbf{51.9} & \underline{59.8} \\
		TAE \cite{li2020learning} &  46.3 & 48.8 & 13.4 & 76.7 & 74.8 & 81.8 & 19.9 & 42.3 & 50.6 & 50.0 & 50.5 \\
		EmoCo~\cite{sun2021emotion} & 51.8 & 42.9 & 22.9 & 79.8 & 77.0 & \underline{85.2} & 23.4 & 42.5 & 55.4 & 49.6 & 53.0 \\ 
	    CLP~\cite{li2023contrastive}  & 44.6 & \underline{58.7} &
		34.7 & 75.9 & \textbf{78.6} & \textbf{86.6} & 20.3 & 44.8 & 56.4 & 42.2 & 54.3 \\
		AAR~\cite{shao2023facial} & 66.3 & 53.9 & 23.7 & 81.5 & 73.6 & 84.2 & 43.8 & 53.8 & 58.2 & 46.5 & 58.5 \\
	\midrule
		\textbf{HiVA (Ours)} & \underline{56.5} & \textbf{59.6} &
		\underline{52.4} & \textbf{82.6} & 77.7 & 80.2 & 40.8 & \underline{47.8} & \textbf{59.3} & 49.9 & \textbf{60.7} \\ 
	\bottomrule
	\end{tabular}
\end{table*}

\begin{figure*}[htb]
	\centering
	\includegraphics[width=1\textwidth]{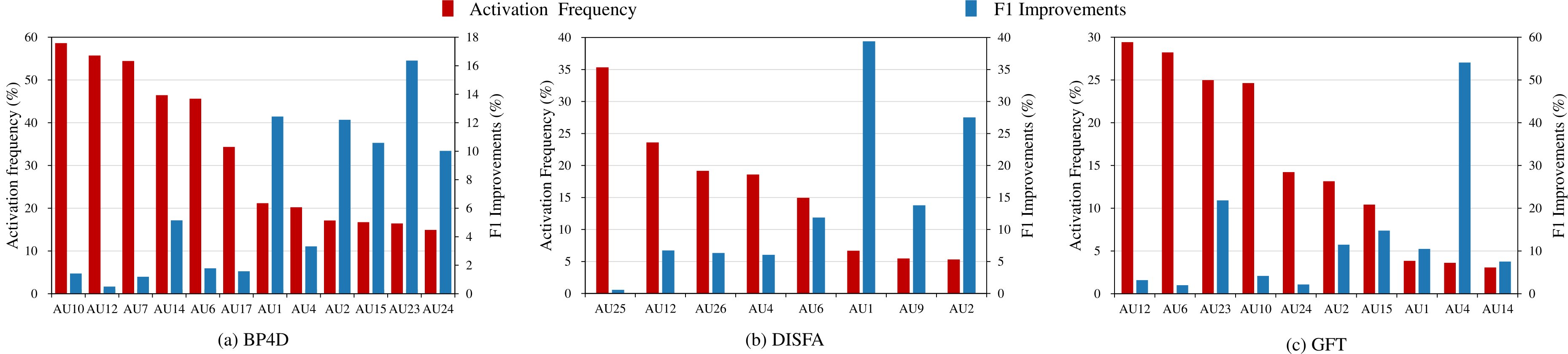}
	\caption{Comparison of per-AU F1 score improvements between the proposed HiVA and its vision-only baseline. HiVA consistently enhances AU detection performance, particularly for infrequently activated AUs. This trend highlights the effectiveness of language-based AU descriptions in improving feature robustness and modeling inter-AU dependencies via dual cross-modal attention mechanisms.
	}
	\label{fig:AU-Text-frequency}
\end{figure*}

\section{Experiments}
\label{sec:experiments}

\textbf{Implementation details and Evaluation protocols}.
The learning rates for the first and second training stages were set to 1e-5 and 1e-6, respectively. For the second stage, the hyperparameter $\lambda$ was empirically set to 1e-5 based on validation performance. All experiments were implemented in PyTorch and conducted on two RTX 3090 GPUs, each equipped with 24 GB of memory. The batch size was set to 128, and the model was trained for 30 epochs until convergence. For model evaluation, the AU-Aware Dynamic Graph Construction module is excluded, and the remaining modules (i.e., components (a), (c), and (d) in Fig.~\ref{fig:framework}) are combined for performance assessment. In the inference phase, pre-trained AU descriptions serve as semantic priors, guiding the vision-language interaction to align visual features with AU-specific textual cues. The F1 score ($F1=\frac{2RP}{R+P}$), where $R$ and $P$ denote recall and precision, respectively, is employed as the primary evaluation metric. The final performance is reported as the average F1 score across all AUs, with results presented in percentage form ($F1 \times 100$). HiVA model comprises 156.08M parameters and requires 19.09 GFLOPs (floating-point operations) per inference. In terms of inference efficiency, HiVA achieves an average runtime of 63.1 ms per inference on an NVIDIA RTX 3090 GPU (24 GB memory) with a batch size of 32. On CPU-based inference (Intel Xeon Platinum 8352V), the model requires 215.8 ms per inference.
 
For network architecture, the Swin Transformer Base model was employed as the visual backbone. Input images were resized to $3 \times 224 \times 224$, and the output feature map from the Swin Transformer Base ($7 \times 7 \times 1024$) was projected to a global feature map of size $7 \times 7 \times 512$ via pointwise convolution. To learn AU-specific features $\mathbf{U}_i \in \mathbb{R}^{512}$ (as shown in Fig.~\ref{fig:figure3}), parallel sub-branches were introduced, each employing a one-layer pointwise convolution to produce intermediate features $\mathbf{F}_i \in \mathbb{R}^{7 \times 7 \times 512}$. For the Semantic Encoding module, we utilized the bert-large-uncased model, fine-tuning only the last two transformer layers to obtain AU description embeddings $\mathbf{H}_i \in \mathbb{R}^{12 \times 512}$. In the Contextualized Modeling module, a 3-layer transformer was applied, where each layer consists of 8 attention heads with a hidden dimension of 512.

\textbf{Data Preparation.}
We evaluate HiVA on three publicly available FACS datasets: BP4D~\cite{zhang2014bp4d}, DISFA~\cite{mavadati2013disfa}, and GFT~\cite{girard2017sayette}. The BP4D dataset consists of approximately 146,000 frames collected from 41 subjects (23 females and 18 males), with annotations for 12 distinct AUs. The DISFA dataset includes roughly 130,000 frames from 27 participants, each annotated for 8 AUs with intensity levels ranging from 0 to 5; following standard practice, frames with intensity scores greater than 1 are treated as positive samples, while those with lower scores are considered negative. Evaluation on BP4D and DISFA is conducted using a subject-independent three-fold cross-validation protocol. The GFT dataset captures spontaneous social interactions among 96 participants grouped into 32 triads, introducing additional challenges due to moderate out-of-plane head movements. Each participant contributes approximately 1,800 frames labeled with 10 AUs. For GFT, we adopt the original train/test splits provided in~\cite{girard2017sayette}, comprising approximately 108,000 training images and 24,600 test images.

\subsection{Comparison with the state-of-the-art methods}
We compare HiVA with the representative AU detection methods, including EAC-Net \cite{li2017eac}, DSIN \cite{corneanu2018deep}, ATF \cite{zhang2018identity}, IdenNet \cite{tu2019idennet}, JAA-Net \cite{shao2018deep}, SRERL \cite{li2019semantic}, SO-Net \cite{yang2020set}, MAL \cite{li2021meta}, UGN~\cite{song2021uncertain},   TAE~\cite{li2020learning}, PIAP~\cite{tang2021piap}, CISNet~\cite{chen2022causal}, AUNet~\cite{yang2023toward}, ME-GraphAU~\cite{luo2022learning}, MGRR-Net~\cite{ge2024mgrr}.
We also compare HiVA with the language-guided AU detection methods, including SEV-Net~\cite{yang2021exploiting}, CLEF~\cite{zhang2023weakly} and VL-FAU~\cite{ge2024towards}.

Among existing state-of-the-art AU detection approaches, methods such as EAC-Net~\cite{li2017eac}, JAA-Net~\cite{shao2018deep}, DSIN~\cite{corneanu2018deep}, ARL~\cite{shao2019facial}, and SRERL~\cite{li2019semantic} primarily rely on manually selected facial landmarks to extract regional facial features, and employ dedicated convolutional branches to learn region-specific AU representations. In contrast, PIAP~\cite{tang2021piap} adopts a pixel-level attention mechanism to localize AU-related regions and suppresses identity-related features by leveraging an auxiliary face recognition model. These methods are evaluated and directly compared with the proposed HiVA framework under identical experimental protocols on the BP4D and DISFA datasets. In addition, AUNet~\cite{yang2023toward} incorporates a stacked hourglass-based face alignment module and models temporal dynamics using a temporal difference network, which enhances static representations with localized motion cues.

Based on the quantitative comparisons presented in Tab.~\ref{tab:bp4d_cross_dataset}, \ref{tab:disfa_cross_dataset}, \ref{tab:gft_cross_dataset} on BP4D, DISFA, GFT datasets, we derive the following key observations: Firstly, The proposed HiVA framework consistently outperforms state-of-the-art vision-based AU detection methods, including Jacob \textit{et al}~\cite{jacob2021facial}, PIAP~\cite{tang2021piap}, ME-GraphAU~\cite{luo2022learning},  MDHR~\cite{wang2024multi}. This consistent advantage empirically validates the effectiveness of augmenting AU representations through language-based auxiliary descriptions, e.g., compared with CISNet~\cite{chen2022causal}, a method that incorporates causal intervention and requires additional facial identity annotations, HiVA achieves superior F1 scores across all benchmark datasets. Moreover, HiVA surpasses ME-GraphAU, which explicitly incorporates multi-dimensional edge features to model task-specific AU dependencies. Notably, although AUNet~\cite{yang2023toward} and MGRR-Net~\cite{ge2024mgrr} demonstrates competitive performance, HiVA distinguishes itself by eliminating the need for auxiliary pre-trained face alignment models, thus improving its practical applicability and ease of deployment. 

Secondly, HiVA also surpasses existing multimodal AU detection methods by a significant margin, e.g., SEV-Net~\cite{yang2021exploiting} integrates AU descriptions using a standard cross-attention mechanism, yet falls short of HiVA’s performance. This improvement can be primarily attributed to two key factors: the construction of semantically enriched, language-based AU representations and the design of a hierarchical vision-language interaction paradigm. 
Further, HiVA demonstrates improvements over VL-FAU~\cite{ge2024towards}, validating the effectiveness of the DDCA and CDCA mechanisms.
The individual contributions of these components will be thoroughly analyzed in Section~\ref{sec:exp_ablation_study}.

Thirdly, HiVA exhibits consistently strong performance across most individual AUs. This is particularly significant given the substantial imbalance in AU activation frequencies, which poses a major challenge to attaining uniformly high performance across all AUs. These results highlight HiVA’s effectiveness in enhancing both the representation and recognition of a wide range of AUs, and they motivate further exploration of its advantages over baseline methods that rely solely on visual cues.

The comparative results w.r.t per-AU's relative improvements are visualized in Fig.~\ref{fig:AU-Text-frequency}. Here, the baseline denotes a vision-only variant of HiVA, in which the language branch is removed. The figure clearly illustrates that HiVA achieves consistent gains across nearly all AUs. Notably, a distinct pattern emerges: the lower the activation frequency of a specific AU, the greater the relative improvement in detection accuracy achieved by HiVA. This trend can be attributed to two primary factors. First, AUs with low activation frequency benefit more substantially from the auxiliary semantic guidance provided by textual descriptions, which supports the learning of more discriminative and robust features. Second, the integration of AU descriptions facilitates the modeling of richer and more reliable inter-AU relationships through HiVA’s dual cross-modal attention mechanisms (DDCA and CDCA). This finding underscores the crucial role of incorporating language-based AU supervision in enhancing the robustness, expressiveness, and generalizability of AU feature representations.

\begin{table*}[t]
	\centering
	\caption{Ablation study on BP4D dataset. \textbf{Bold} and \underline{underline} denote the best and second best results, respectively.}
	\begin{tabular}{c|c|c|c|c|c|c|c|c|c|c|c|c|c}
			\toprule
		\textbf{Methods/AU} & AU1 & AU2 & AU4 & AU6 & AU7 & AU10 & AU12 & AU14 & AU15 & AU17 & AU23 & AU24 & \textbf{AVE} \\
		\midrule
		HiVA (w/o AUG, w/o $L_{\text{diff}}$) & 50.9 & 47.5 & 62.1 & 78.2 & 78.8 & 83.6 & 87.9 & 66.0 & 52.6 & 61.5 & 48.0 & 53.0 & 64.2 \\
		HiVA (w/o AUG) & 52.6 & 49.0 & 62.9 & 78.3 & 78.3 & 83.6 & \underline{88.6} & 65.8 & 54.9 & 61.1 & 49.7 & 53.3 & 64.9 \\
		HiVA (w/o $L_{\text{diff}}$) & 53.2 & 48.1 & \underline{63.0} & 78.2 & \underline{79.2} & 83.7 & 88.5 & \underline{67.1} & 54.9 & 61.5 & \underline{50.8} & \underline{55.2} & \underline{65.4} \\
     \midrule
			
			\textbf{HiVA (Ours)} & \textbf{54.1} & \textbf{49.7} & \textbf{63.3} & \textbf{79.3} & \textbf{79.8} & \textbf{84.5} & \textbf{88.8} & \textbf{68.5} & \textbf{57.0} & \textbf{62.6} & \textbf{53.1} & \textbf{56.8} & \textbf{66.5} \\
		\midrule
		VLU & 48.1 & 44.3 & 61.3 & 77.9 & 78.9 & 83.3 & 88.4 & 65.1 & 51.5 & 61.6 & 45.6 & 51.6 & 63.1 \\

		HiVA (w/o DDCA) & 51.4 & 48.8 & \underline{63.0} & 78.5 & \underline{79.2} & 83.4 & 88.4 & 65.8 & 54.8 & 60.4 & 50.2 & 53.4 & 64.8 \\
		HiVA (w/o CDCA) & \underline{53.9} & 48.7 & 62.7 & \underline{78.6} & 79.1 & \underline{84.1} & 88.5 & \underline{67.1} & \underline{55.3} & \underline{61.9} & 50.2 & 55.0 & \underline{65.4} \\
		HiVA (w/o CDCA, w/o DDCA) & 52.5 & \underline{49.6} & 62.8 & 78.1 & 78.1 & 83.3 & 88.4 & 64.4 & 54.5 & 61.0 & 48.7 & 51.4 & 64.4 \\
		\bottomrule
	\end{tabular}
	\label{tab:ablation_bp4d}
\end{table*}

\begin{table*}[t]
	\centering
	\caption{Ablation study on DISFA dataset. \textbf{Bold} and \underline{underline} denote the best and second best results, respectively.}
	\begin{tabular}{c|c|c|c|c|c|c|c|c|c}
		\toprule
		\textbf{Methods/AU} & AU1 & AU2 & AU4 & AU6 & AU9 & AU12 & AU25 & AU26 & \textbf{AVE} \\
		\toprule
		HiVA (w/o AUG, w/o $L_{\text{diff}}$) & 47.4 & 49.7 & 73.7 & 51.0 & 54.8 & 73.5 & \textbf{95.5} & \underline{67.8} & 64.2 \\
		HiVA (w/o AUG) & 51.2 & 51.1 & 73.1 & 50.6 & 53.6 & \textbf{76.7} & 94.8 & 65.6 & 64.6 \\
		HiVA (w/o $L_{\text{diff}}$) & 45.9 & 46.6 & \textbf{78.5} & \textbf{53.9} & \underline{59.2} & 72.7 & 95.1 & 66.0 & 64.6 \\
    \midrule
	\textbf{HiVA (Ours)}  &  \textbf{60.6} & \textbf{58.4} & \underline{75.4} & 51.0 & \textbf{61.2} & \underline{74.8} & 93.9 & 63.8 & \textbf{67.4} \\
	\midrule
		Baseline & 43.5 & 45.8 & 71.1 & 45.6 & 53.8 & 70.1 & 93.4 & 60.0 & 60.4 \\
		HiVA (w/o DDCA) & 52.4 & 50.8 & 72.0 & 50.1 & 54.3 & 72.7 & 95.0 & 67.3 & 64.3 \\
		HiVA (w/o CDCA) & \underline{53.5} & \underline{55.5} & 72.9 & \underline{51.6} & 55.9 & 73.6 & 94.6 & 67.1 & \underline{65.5} \\
		HiVA (w/o CDCA, w/o DDCA) & 47.3 & 46.5 & 70.8 & 49.2 & 51.9 & 71.6 & \underline{95.3} & \textbf{68.0} & 62.6 \\
		\bottomrule
	\end{tabular}
	\label{tab:ablation_disfa}
\end{table*}

\begin{table*}[t]
	\centering
	\caption{Ablation study on GFT dataset. \textbf{Bold} and \underline{underline} denote the best and second best results, respectively.}
	\begin{tabular}{c|c|c|c|c|c|c|c|c|c|c|c}
		\toprule
		\textbf{Methods/AU} & AU1 & AU2 & AU4 & AU6 & AU10 & AU12 & AU14 & AU15 & AU23 & AU24 & \textbf{AVE} \\
		\midrule
		HiVA (w/o AUG, w/o $L_{\text{diff}}$) & 50.8 & 57.1 & 44.8 & 80.2 & 75.9 & 77.3 & 39.3 & 45.3 & 55.0 & \underline{50.6} & 57.6 \\
		HiVA (w/o AUG) & 50.9 & 56.8 & 47.5 & 81.3 & 76.2 & 78.0 & \underline{40.7} & 46.4 & 56.6 & 51.2 & 58.6 \\
		HiVA (w/o $L_{\text{diff}}$) & 51.0 & \underline{58.7} & \textbf{54.7} & 80.4 & 77.0 & 80.1 & 38.0 & \underline{48.4} & \underline{58.9} & 47.7 & 59.5 \\
	\midrule
		\textbf{HiVA (Ours)} & \textbf{56.5} & \textbf{59.6} & \underline{52.4} & 82.6 & \underline{77.7} & \textbf{80.2} & \textbf{40.8} & 47.7 & \textbf{59.3} & 49.9 & \textbf{60.7} \\
		\midrule
		Baseline & 51.1 & 53.5 & 34.0 & 81.0 & 74.6 & 77.7 & 37.9 & 41.6 & 48.7 & 48.8 & 54.9 \\
		HiVA (w/o CDCA, w/o DDCA) & 50.6 & 57.3 & 36.7 & \textbf{83.3} & 75.7 & 78.5 & 37.3 & 45.5 & 55.8 & 49.7 & 57.0 \\
		HiVA (w/o DDCA) & 53.6 & 57.3 & 43.6 & 81.9 & \textbf{79.0} & \underline{79.9} & 39.9 & 45.1 & 57.2 & 49.8 & 58.7 \\
		HiVA (w/o CDCA) & \underline{54.3} & 56.8 & 41.7 & \underline{83.2} & 77.3 & 78.1 & 46.3 & \textbf{49.4} & 53.2 & \textbf{50.7} & \underline{59.7} \\
	\bottomrule
	\end{tabular}
	\label{tab:ablation_gft}
\end{table*}

\begin{figure}[htb]
	\centering
	\includegraphics[width=0.48\textwidth]{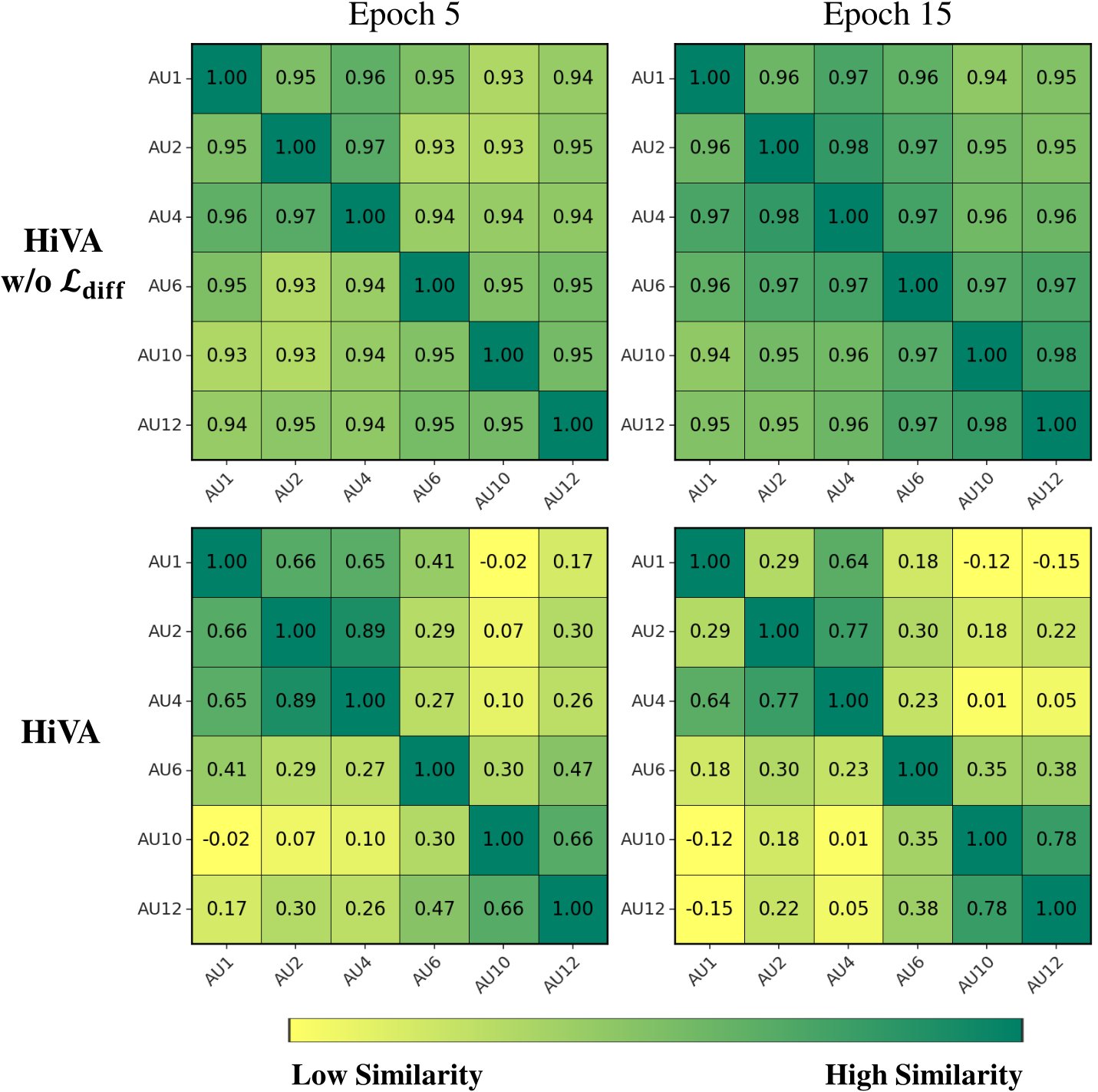}
	\caption{Visualization of the cosine similarity matrices among textual AU embeddings at different training epochs for HiVA (w/o $\mathcal{L}_{\text{diff}}$) (top) and HiVA (bottom). HiVA progressively learns more discriminative and diverse AU embeddings, while the variant without $\mathcal{L}_{\text{diff}}$ shows persistently high similarity across different AUs, indicating limited semantic differentiation.
	}
	\label{fig:diff_ablation_study}
\end{figure}

\begin{figure*}[htb]
	\centering
	\includegraphics[width=0.70\textwidth]{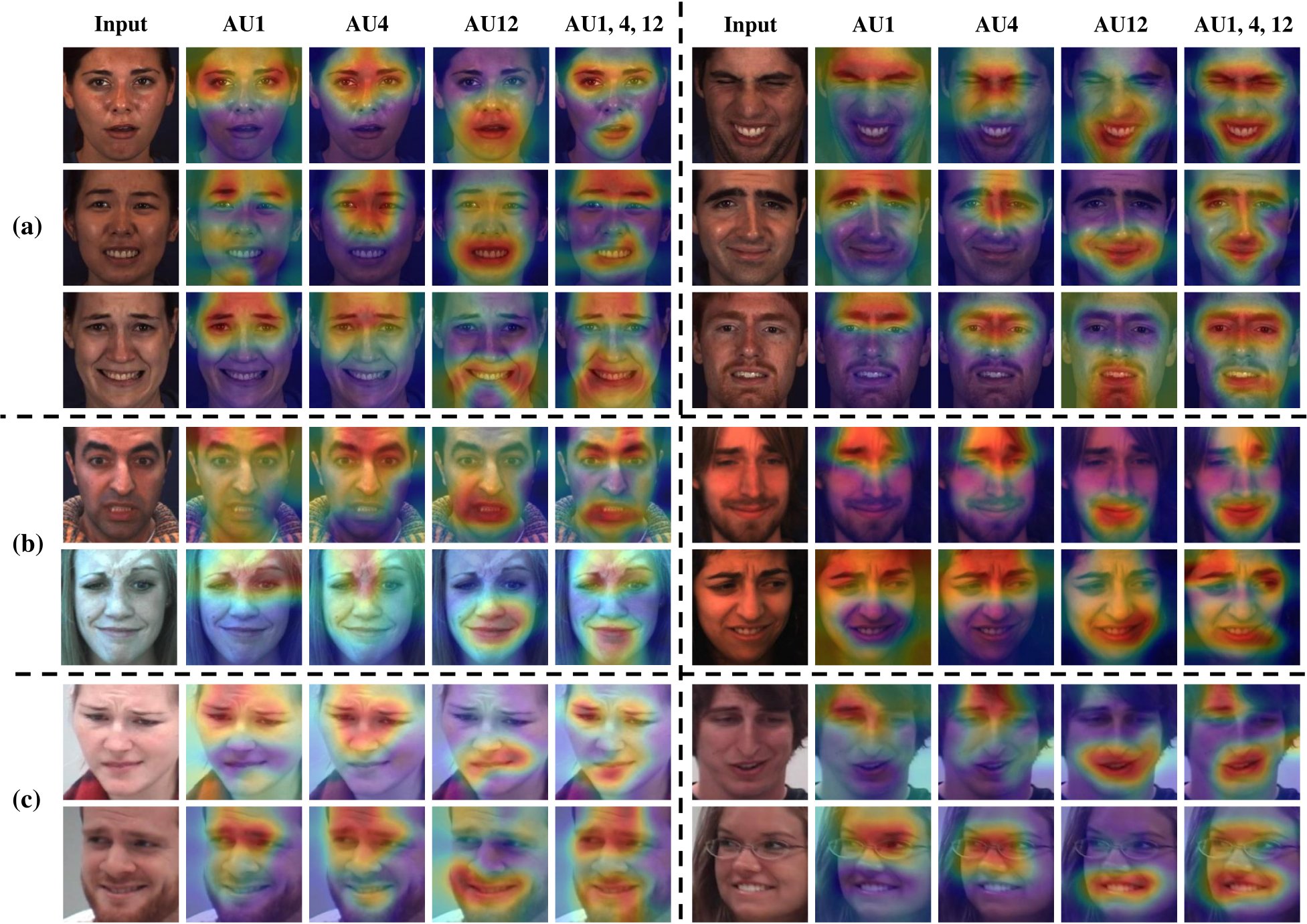}
	\caption{Visualization of the learned cross-modality attention maps for several AUs (as examples) of different subjects on BP4D (a), DISFA (b), and GFT (c) datasets. The first three attention maps for each example are derived from DDCA. The final composite attention map is generated by CDCA.
	}
	\label{fig:attention_map}
\end{figure*}

\subsection{Analysis of Each Component in HiVA}
\label{sec:exp_ablation_study}
We conduct ablation studies to investigate the contributions of the following two components in HiVA: (1)  GPT4-augmented AU descriptions and (2) the regularization loss $\mathcal{L}_{\text{diff}}$. 
Besides, we also investigate the contributions of DDCA, CDCA and thus compare HiVA with its counterparts without one or two of them. We show the comparison in Tab.~\ref{tab:ablation_bp4d}, Tab.~\ref{tab:ablation_disfa}, Tab.~\ref{tab:ablation_gft}.

\textbf{Quantitative analysis.} Firstly, we assess the contributions of the GPT-4-augmented AU descriptions and the regularization loss term $\mathcal{L}_{\text{diff}}$ to the overall performance of the HiVA framework. As shown in Tables~\ref{tab:bp4d_cross_dataset}, \ref{tab:disfa_cross_dataset}, and \ref{tab:gft_cross_dataset}, the variant “HiVA (w/o AUG, w/o $\mathcal{L}{\text{diff}}$)” represents the model trained without these two components. From the comparative results across all three datasets, the following key observations can be drawn: (1) The integration of GPT-4-augmented AU descriptions consistently improves performance across the BP4D, DISFA, and GFT datasets. This enhancement can be attributed to the increased diversity and richness of the textual AU descriptions provided by GPT-4, which introduces varied linguistic cues beneficial for learning robust AU representations. (2) The regularization loss $\mathcal{L}_{\text{diff}}$ proves to be a crucial component for HiVA, as it encourages the model to capture subtle distinctions among different AU descriptions, thus facilitating fine-grained discriminative feature learning. To illustrate this, Fig.~\ref{fig:diff_ablation_study} visualizes the cosine similarities between selected AU embeddings at different training epochs for both the complete HiVA model and the variant without $\mathcal{L}_{\text{diff}}$. The full model shows progressively improved differentiation between AUs over time, while the variant without the regularization term exhibits persistently high inter-AU similarity, indicating reduced discriminative capacity.
(3) The highest overall performance is achieved when both GPT-4-augmented descriptions and the $\mathcal{L}_{\text{diff}}$ loss are jointly employed. This suggests that the two components are complementary: the enriched semantic input from GPT-4 provides diverse contextual signals, while the regularization term enforces discriminative alignment in the learned embedding space. Together, they contribute synergistically to the advancement of AU detection accuracy.

Secondly, we evaluate the effectiveness of the two key components in the proposed hierarchical vision-language interaction module: DDCA and CDCA. In Tables~\ref{tab:bp4d_cross_dataset}, \ref{tab:disfa_cross_dataset}, and \ref{tab:gft_cross_dataset}, the “Baseline” variant refers to a vision-only model that excludes the language branch and utilizes only the visual backbone with global and local features for AU detection. From the results, we derive the following key findings:
From the comprehensive comparisons in the tables, we can conclude that: (1) Both DDCA and CDCA play essential and complementary roles in improving HiVA’s AU detection capabilities. DDCA provides fine-grained alignment between local visual features and AU-specific semantic cues, while CDCA captures global contextual dependencies between modalities, facilitating more comprehensive understanding of facial actions. Their joint use significantly enhances the model’s ability to represent both localized and holistic AU-related information.
(2) Among the two, DDCA is slightly more influential overall. This is likely due to its ability to directly associate each AU-specific visual representation with its corresponding textual description at a fine-grained level, which is particularly effective for capturing subtle, localized facial muscle movements for AU detection.
(3) The integration of both components enables HiVA to perform fine-grained recognition and holistic semantic reasoning, resulting in more robust and generalizable AU detection. Their synergistic effect is especially beneficial for detecting low-frequency AUs and for maintaining performance on challenging datasets such as GFT.

\begin{figure}[htb]
	\centering
	\includegraphics[width=0.45\textwidth]{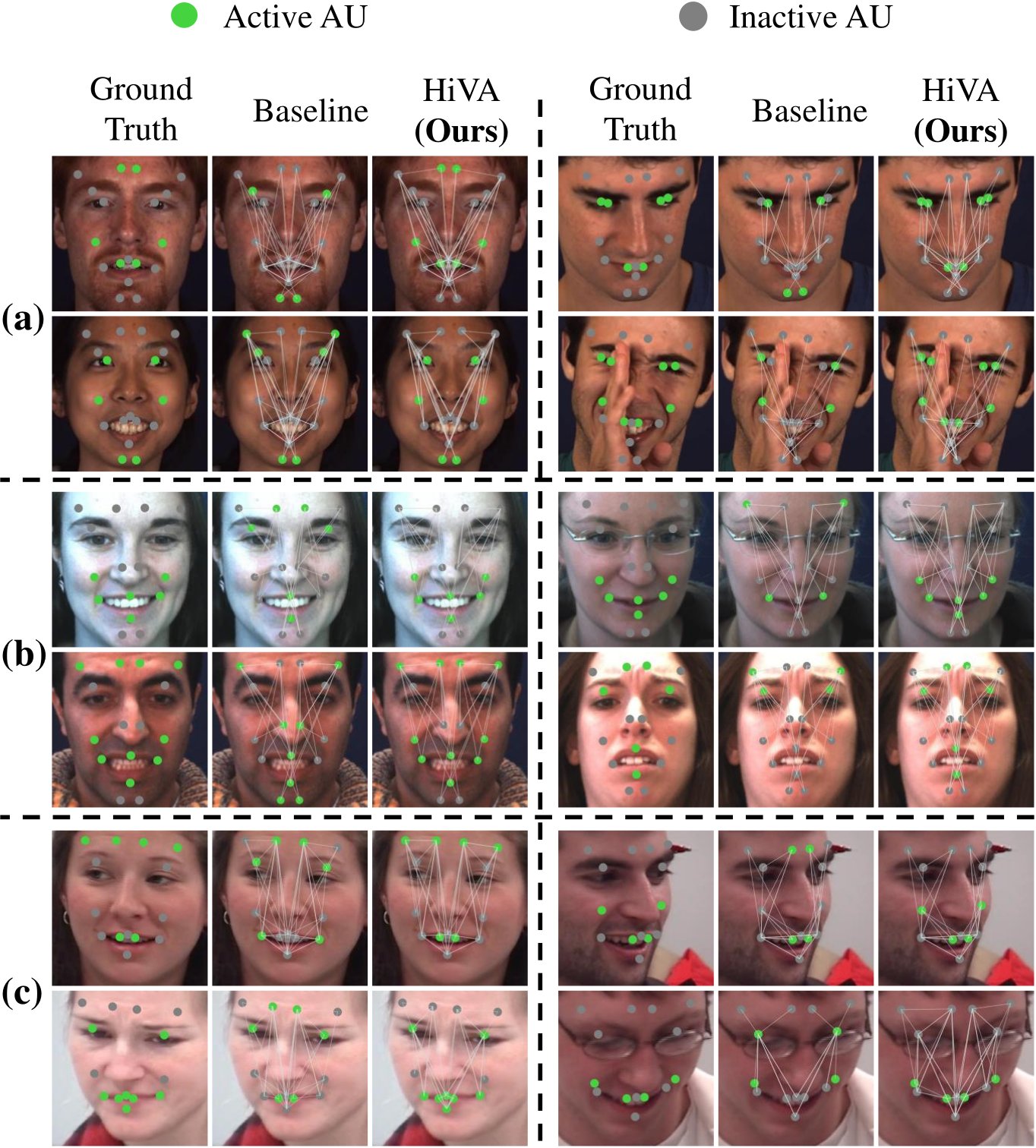}
	\caption{ Visualization of the learned graph structures on the BP4D (a), DISFA (b), and GFT (c) datasets. Each node is connected to its \textit{K} nearest neighbors, with nodes corresponding to activated AUs typically exhibiting a higher number of connections compared to those associated with inactivated AUs. HiVA-generated graphs tend to show more structured and meaningful inter-AU connections.
	}
	\vspace{-2em}
	\label{fig:graph_visualize}
\end{figure}

\textbf{Quantitative analysis.} 
To illustrate how HiVA performs fine-grained and global semantic interaction between facial regions and AU descriptions, we visualize the learned cross-modality attention maps w.r.t DDCA and CDCA for several AUs. These visualizations are generated during inference and rely solely on fixed textual representations derived from the pre-trained model, independent of both the original and generated AU descriptions. The results are shown in Fig.~\ref{fig:attention_map}. For each subject and dataset, the first three attention maps show the DDCA results, where each attention map corresponds to one specific AU. The last composite attention map is derived from the CDCA, which aggregates global contextual information across the three AU descriptions. These attention maps highlight the HiVA’s ability to localize and attend to semantically meaningful facial regions relevant to specific AUs, e.g., AU1 corresponds to movements near the inner eyebrow region, which the attention maps accurately emphasize. The consistent and interpretable localization patterns across subjects and datasets validate the effectiveness of DDCA and CDCA in HiVA.

To illustrate how HiVA captures inter-AU relationships through graph-based modeling, and how these relationships vary depending on AU activation status and the employed method, we visualize the learned AU graph structures in Fig.~\ref{fig:graph_visualize}. 

Obviously, the connection pattern of the Baseline model is sparser and exhibits weaker correlations with the target AUs. It fails to integrate contextual information between AUs, introducing noise that disrupts the AU recognition process, as seen in the incorrect predictions of AU1 and AU4 in the top-left row (Fig.~\ref{fig:graph_visualize} (b)) and AU4 in the bottom-right row (Fig.~\ref{fig:graph_visualize} (c)). In contrast, HiVA's edge connection pattern is denser, capturing complex semantic relationships between AUs that the Baseline model overlooks, thus improving AU prediction accuracy. For example, the meaningful connections between AU6 and AU10 (bottom-right row in Fig.~\ref{fig:graph_visualize} (c)), the connections between AU6 and AU12 (top-left row in Fig.~\ref{fig:graph_visualize} (b)) are more clearly defined in HiVA.
Conclusively, HiVA-generated graphs exhibit more structured and meaningful inter-AU connections, benefiting from the integration of semantic cues from textual AU descriptions.

\section{Conclusion}
\label{sec:conclusion}
Within this manuscript we introduce a novel framework, Hierarchical Vision-Language Attention for AU Understanding (HiVA), designed to enhance AU detection through hierarchical and collaborative vision-language interaction. To Address the limitation of scarce textual knowledge, HiVA leverages a large language model to generate diverse AU descriptions. The core of HiVA lies in its hierarchical cross-modal attention mechanisms, which include disentangled dual cross-attention between AU-specific visual features and their corresponding textual descriptions, and contextual dual cross-attention between global visual features and all AU descriptions. Experimental results consistently show HiVA's ability to improve AU detection performance and reveal semantically meaningful activation patterns. In future work, we intend to extend this framework towards cross-domain AU detection by further exploiting textual AU descriptions.

\section{Acknowledgment}
This research was partially supported by the National Natural Science Foundation of China (62576099, 62276134, 62506070, 62302219), Start-up Research Fund of Southeast University (RF1028625087), RIE2020 AME Programmatic Fund, Singapore (No. A20G8b0102), the A*STAR Prenatal / Early Childhood Grant (No. H22P0M0002), the Natural Science Foundation of Jiangsu Province (Grants No BK20251348), the Research Project of Humanities and Social Sciences of the Ministry of Education (No. 25YJCZH372), the Fundamental Research Funds for the Central Universities (30924010918, 2242025S30059), Joint Open Project of Southeast University-Jiangsu Province Hospital, and the Big Data Computing Center of Southeast University.

	\bibliographystyle{IEEEtran}
	\bibliography{egbib}
	
\begin{IEEEbiography}[{\includegraphics[width=1in,height=1.25in,clip,keepaspectratio]{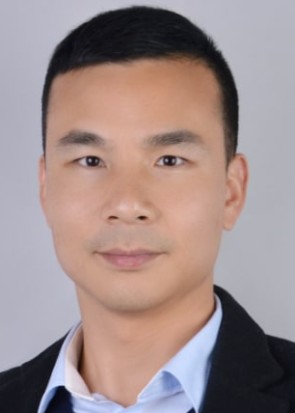}}]
	{Yong Li} is currently an Associate Professor with the School of Computer Science and Engineering, Southeast University, China.  His research interests include deep learning, human-centered affective computing. His research results have been expounded in more than 50 publications at prestigious journals and prominent conferences, such as IEEE TPAMI, IEEE TIP, IEEE TAC, IEEE TMM, NeurIPS, CVPR, ICCV.  For more information, please visit his personal website: https://mysee1989.github.io/english/.
\end{IEEEbiography}

\begin{IEEEbiography}[{\includegraphics[width=1in,height=1.25in,clip,keepaspectratio]{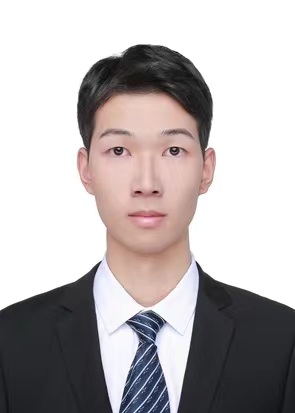}}]
	{Yi Ren} received the bachelor's degree in Software Engineering from Zhengzhou University, Zhengzhou, China, in 2023. His research interests include deep learning, affective computing, and multimodal learning.
\end{IEEEbiography}

\begin{IEEEbiography}[{\includegraphics[width=1in,height=1.25in,clip,keepaspectratio]{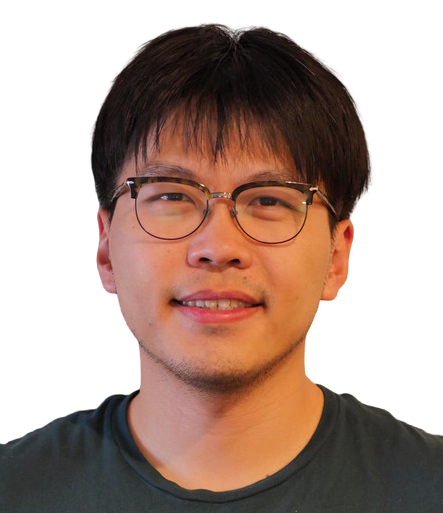}}]
	{Yizhe Zhang} is an associate professor in the School of Computer Science and Engineering at Nanjing University of Science and Technology. Dr. Zhang earned his Ph.D. from the University of Notre Dame, USA. After that, he worked as a researcher in Qualcomm AI research. His primary research focus is on medical (and biological) image analysis, machine learning and algorithm design.
\end{IEEEbiography}

\begin{IEEEbiography}[{\includegraphics[width=1in,height=1.25in,clip,keepaspectratio]{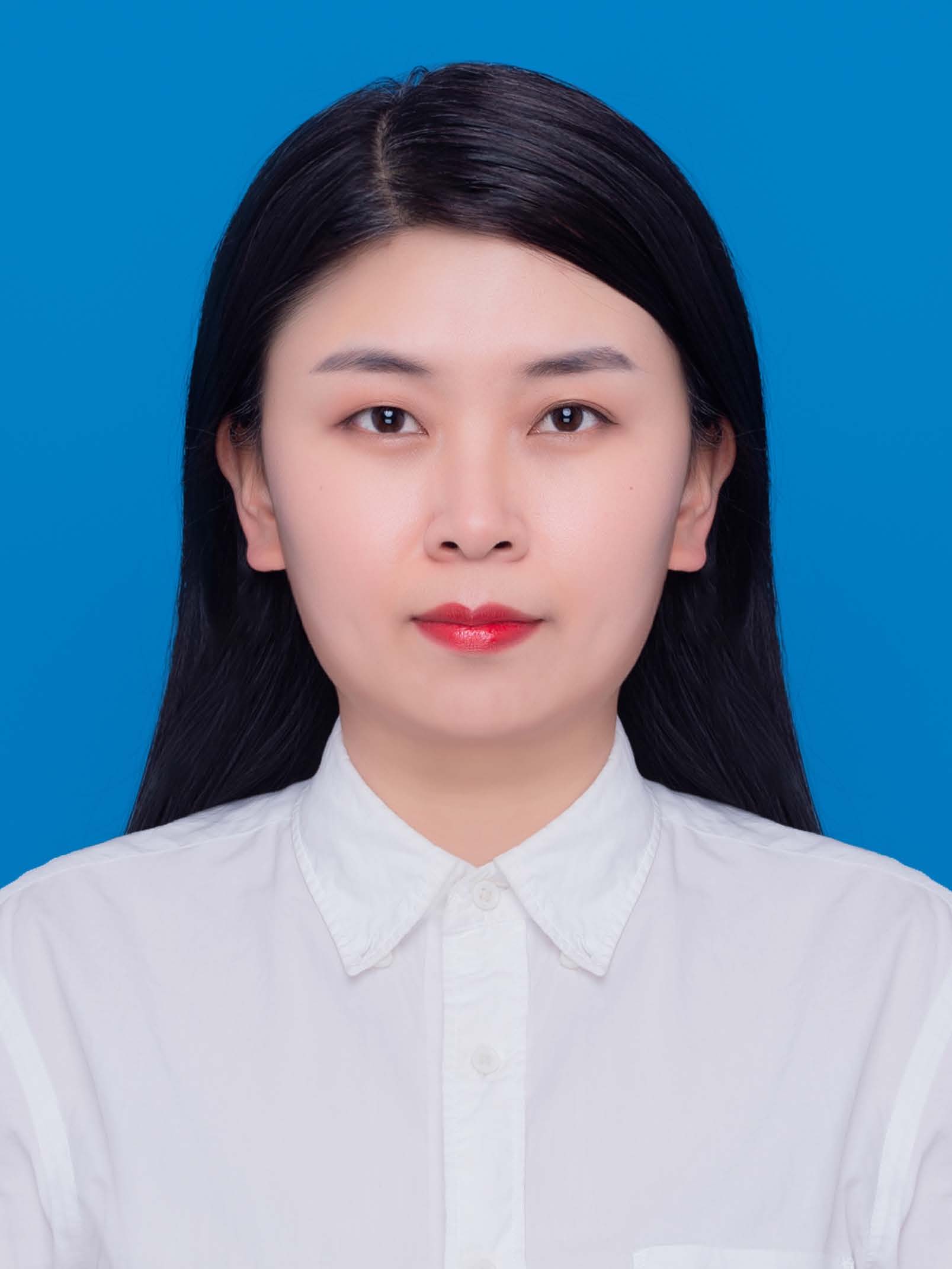}}]
	{Wenhua Zhang} received the B.S. degree from North University of China, Taiyuan, China, in 2015, and the Ph.D. degree from Xidian University, Xi’an, China, in 2021. She is currently an associate professor in the School of Computer Science and Engineering, Nanjing University of Science and Technology, Nanjing, China. Her research interests include object tracking and image processing.
\end{IEEEbiography}

\begin{IEEEbiography}[{\includegraphics[width=1in,height=1.25in,clip,keepaspectratio]{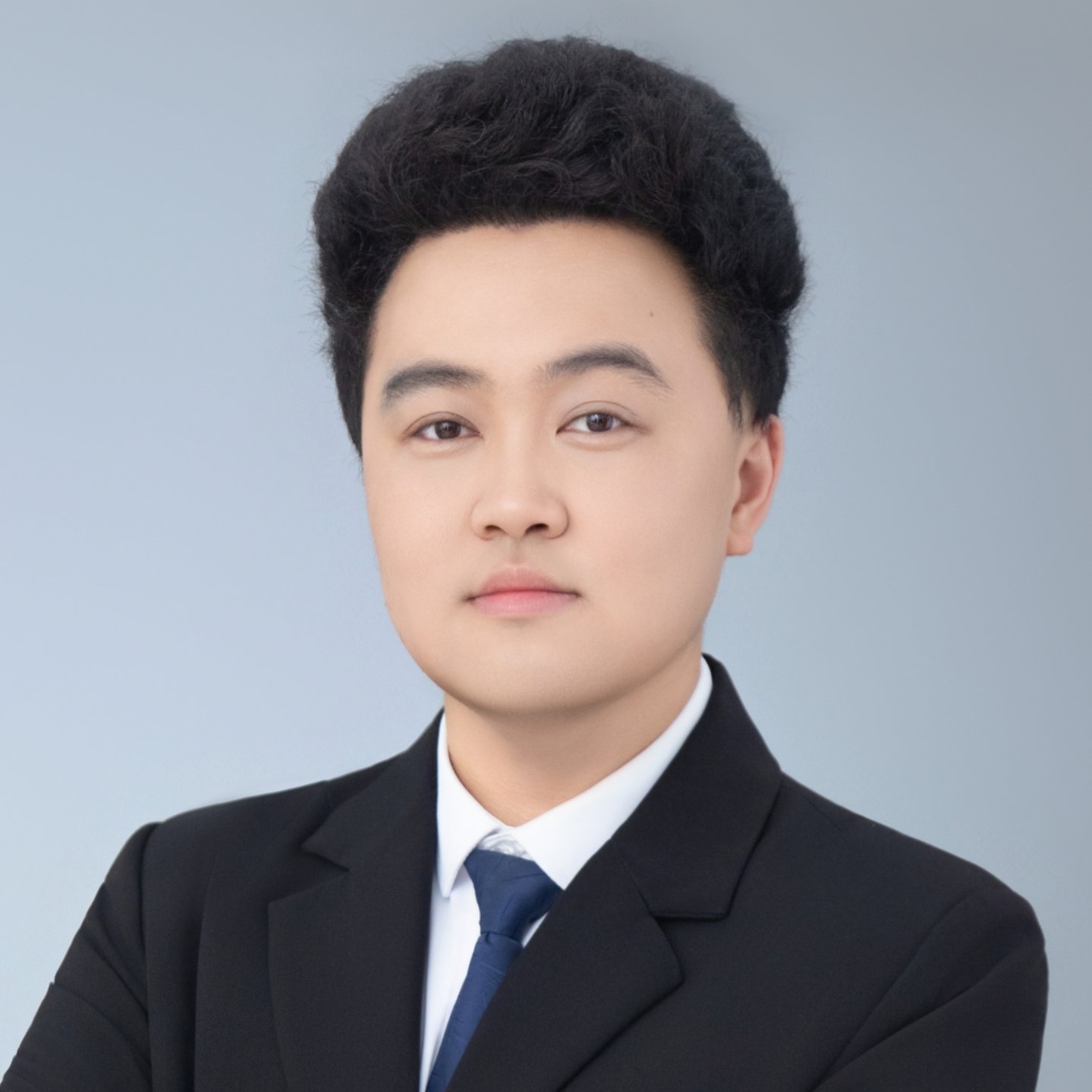}}]
	{Tianyi Zhang} is an associate professor at the School of Biological Sciences and Medical Engineering, Southeast University. Before that, he worked as a postdoc researcher at Vrije Universiteit Amsterdam. He got his PhD degree in Delft University of Technology. He was also associated with the Distributed \& Interactive Systems (DIS) group at the national research institute for mathematics and computer science in the Netherlands (CWI). His research interests lie in affective computing and personality recognition.
\end{IEEEbiography}

\begin{IEEEbiography}[{\includegraphics[width=1in,height=1.25in,clip,keepaspectratio]{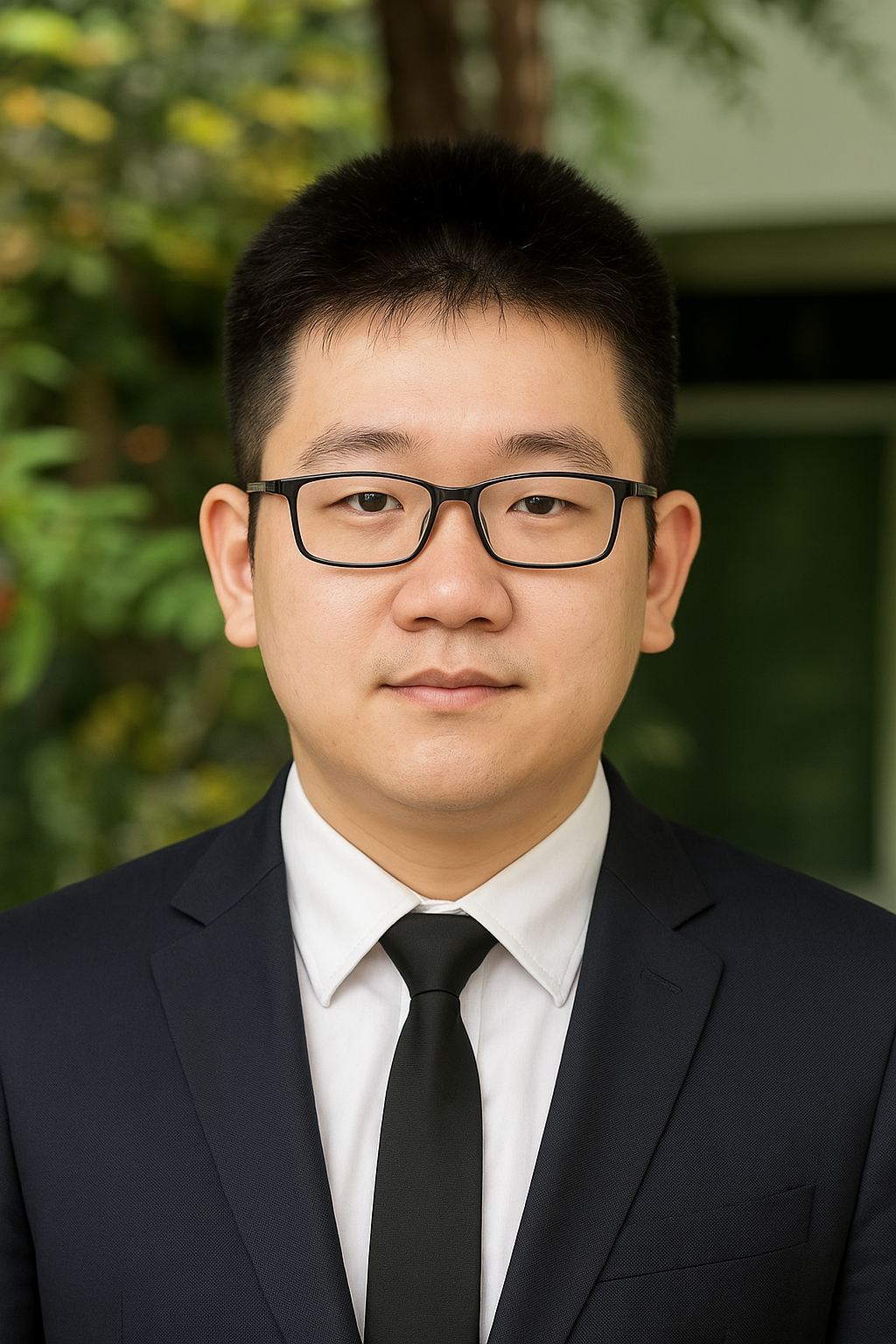}}]
	{Muyun Jiang}  received a bachelor's degree in electrical engineering from the Beijing Institute of Technology, Beijing, China, and a master's degree from Nanyang Technological University (NTU), Singapore. He is currently pursuing his doctoral degree at the College of Computing and Data Science (CCDS) at Nanyang Technological University (NTU), with an interest in Deep Learning for Brain-Computer Interface, EEG decoding, and Covert Speech Decoding.
\end{IEEEbiography}

\begin{IEEEbiography}[{\includegraphics[width=1in,height=1.25in,clip,keepaspectratio]{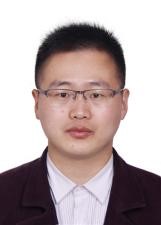}}]
	{Guo-Sen Xie} received the Ph.D. degree from the National Laboratory of Pattern Recognition, Institute of Automation, Chinese Academy of Sciences, Beijing, China, in 2016. He is a Professor at the School of Computer Science and Engineering, Nanjing University of Science and Technology, China. He has published over 70 refereed papers in journals and conferences, including IEEE TPAMI, IJCV, IEEE T-NNLS, IEEE T-IP, IEEE T-CSVT, IEEE T-MM, Pattern Recognition, NeurIPS, CVPR, ICCV, ECCV, AAAI, IJCAI, and ACM MM. He received the Best Student Paper Award of MMM 2016. He is an Associate Editor of IEEE T-IP and Pattern Recognition Journals, and Area Chairs of several international conferences, such as ICLR. His research interests include computer vision and machine learning.
\end{IEEEbiography}

\begin{IEEEbiography}
	[{\includegraphics[width=1in,height=1.25in,clip,keepaspectratio]{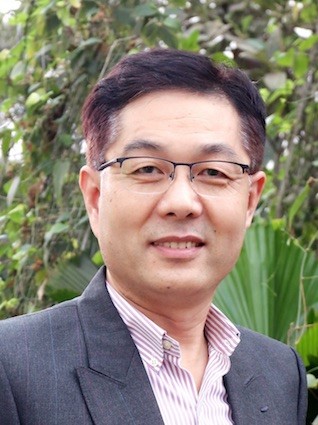}}]{Cuntai Guan}(Fellow, IEEE) received his Ph.D. degree from Southeast University, China, in 1993. He is a President’s Chair Professor in the School of Computer Science and Engineering, Nanyang Technological University, Singapore. He is the Director of the Artificial Intelligence Research Institute, Director of the Centre for Brain-Computing Research, and Co-Director of S-Lab for Advanced Intelligence. His research interests include brain-computer interfaces, machine learning, medical signal and image processing, artificial intelligence, and neural and cognitive rehabilitation. He is a recipient of the Annual BCI Research Award (first prize), King Salman Award for Disability Research, IES Prestigious Engineering Achievement Award, Achiever of the Year (Research) Award, and Finalist of President Technology Award. He is also an elected Fellow of the US National Academy of Inventors (NAI), the Academy of Engineering Singapore (SAEng), and the American Institute for Medical and Biological Engineering (AIMBE).
\end{IEEEbiography}

\end{document}